\documentclass{article}

\PassOptionsToPackage{numbers, compress}{natbib}
 \usepackage[preprint]{neurips_2026}

\usepackage[utf8]{inputenc} 
\usepackage[T1]{fontenc}    
\usepackage{hyperref}       
\usepackage{url}            
\usepackage{booktabs}       
\usepackage{amsfonts}       
\usepackage{nicefrac}       
\usepackage{microtype}      
\usepackage{xcolor}         

\usepackage{subcaption}
\usepackage{makecell}
\usepackage{multirow}
\usepackage{tikz}

\usepackage{amsmath,amssymb,amsfonts}
\usepackage{cleveref}
\DeclareMathOperator*{\argmin}{arg\,min}
\title{SimPersona: Learning Discrete Buyer Personas from
Raw Clickstreams for Grounded E-Commerce Agents}

\author{%
Zahra Zanjani Foumani, Alberto Castelo, Shuang Xie, Ted Chaiwachirasak, Han Li, \\
  \textbf{Lingyun Wang}\thanks{Corresponding Author: lingyun.wang@shopify.com} \\
  Shopify\\
  Bellevue, Washington, USA \\
}

\begin{document}

\maketitle

\begin{abstract}

Large language model (LLM)-based web agents can navigate live storefronts, yet they often collapse to a single "average buyer" policy, failing to capture the heterogeneous and distributional nature of real buyer populations. Existing personalization methods rely on hand-crafted prompt-based personas that are brittle, difficult to scale, context-inefficient, and unable to faithfully represent population-level behavior. We introduce SimPersona, a novel framework that learns discrete buyer types from historical traffic and exposes them to LLM-based web agents as compact persona tokens. Given raw clickstreams, a behavior-aware vector-quantized variational autoencoder (VQ-VAE) induces a discrete buyer-type space that captures the statistical structure of real buyer behavior and merchant-specific buyer population distributions. To provide behavior-specific guidance to LLM-based web agents, \textsc{SimPersona} maps each learned buyer type to a dedicated persona token in the LLM agent vocabulary and fine-tunes the agent with these tokens on real browsing traces. At inference, each synthetic buyer is assigned to a learned buyer type with a single encoder forward pass, requiring no retraining or store-specific prompt engineering. For population-level simulation, \textsc{SimPersona} samples buyer types from each merchant's empirical distribution over the learned VQ-VAE codebook and instantiates agents with the corresponding persona tokens, preserving merchant-specific buyer population distributions. Evaluated on $8.37$M buyers across $42$ held-out live storefronts, \textsc{SimPersona} achieves $78\%$ conversion-rate alignment with real buyers, exhibits interpretable behavioral variation across buyer types, and outperforms a baseline with $8\times$ more parameters on goal-oriented shopping tasks. We further release an open-source data pipeline that converts raw e-commerce event logs into buyer representations and agent-training traces.

\end{abstract}

\section{Introduction}
\label{sec: intro}
\begin{figure}[t]
  \centering
  \includegraphics[width=\textwidth]{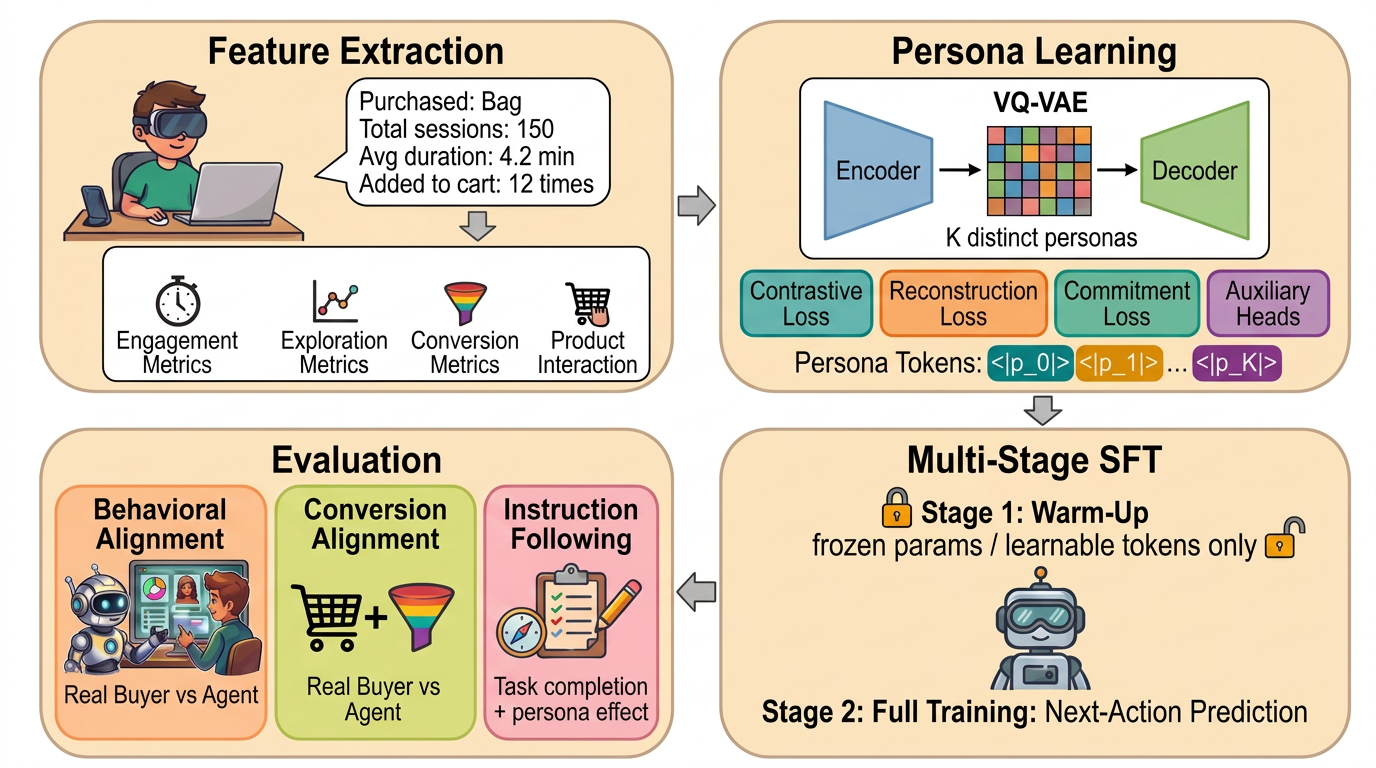}
  \caption{\textsc{SimPersona} framework overview. \textbf{Top-left}: behavioral features and product embeddings are extracted from raw clickstreams. \textbf{Top-right}: a behavior-aware VQ-VAE maps each buyer to one of $K$ persona tokens. \textbf{Bottom-right}: two-stage SFT grounds the tokens in the LLM; first token warm-up (backbone frozen), then full fine-tuning. \textbf{Bottom-left}: evaluation on unseen storefronts across behavioral alignment, conversion alignment, and instruction following.} 
  \label{fig: framework}
\end{figure}

Simulating realistic human shopping behavior has direct commercial impacts where even marginal improvements in buyer modeling translate into measurable gains in recommendation~\citet{ni2018perceive,shi2025personax}, storefront evaluation~\citet{lu2025uxagent}, or synthetic A/B testing~\citet{wang2025agentab}. LLM-based web agents have recently made such simulations feasible at scale, navigating live storefronts and executing complete shopping flows from search to checkout \citet{deng2023mind2web,zhou2024webarena,gur2023realworld}. Yet, while these agents master the \emph{mechanics} of web interaction, they learn a single population-level policy that produces an ``average buyer'': they know \emph{how} to shop, but not \emph{who} they are shopping as.
That is, steering an agent to reflect a specific buyer segment, the \emph{persona problem}, remains the core challenge, and reconstructing the \emph{distribution} of buyer types that collectively defines a store’s traffic remains open~\citet{wang2025opera,ma2025canllm,li2025behavioral}. 

The standard approach to the persona problem is to tell the agent who to be via prompts where memory modules~\citet{park2023generative,park2024generative1000}, purchase-history summaries~\citet{mao2024bought,shi2025personax}, or role-playing profiles~\citet{shao2023characterllm,chuang2024beyond} are leveraged for steering. However, prompt-based conditioning is brittle: behavior shifts with wording~\citet{lee2025prompt}, demographic role-playing fails to match real responses~\citet{chuang2024beyond}, and the best prompt-based shopping agents reproduce only $11.9\%$ of real buyer actions~\citet{ma2025canllm}. Text-based personas are also expensive (requiring auxiliary LLM calls and consuming context at every step) and limited in expressiveness, since hand-crafted templates cannot capture the full diversity of real behavior. Crucially, they do not solve the \emph{distribution} problem, because personas are crafted individually rather than learned from the full population, they carry no notion of how many buyers of each type exist, yet realistic simulation demands not just individual personas but the correct \emph{mix} of buyer types that defines a store's traffic. Recent RL-based agents~\citet{fan2025customerr1,lu2025shopr1} face the same limitation, as their personas also derive from LLM-generated text profiles.

The user-modeling literature takes a data-driven approach, learning structured buyer representations directly from raw clickstreams~\citet{yang2023trace,ni2018perceive,zheng2020deep,kim2017detecting,wang2016unsupervised}. These methods reveal that shopping behavior has rich latent structure that textual profiles cannot express, and in principle can capture population-level variation. While providing high expressivity, they are designed for offline user modeling; their representations are used for tasks such as prediction, recommendation, and clustering, not for driving agent behavior in closed-loop environments.
User modeling solves persona \emph{discovery} but not persona \emph{grounding}: it captures who a buyer is without teaching an agent how to act on the provided information.

We propose \textsc{SimPersona} (\Cref{fig: framework}), a framework that jointly addresses persona discovery, persona grounding, and population distribution learning by representing buyer behavior as \emph{discrete persona tokens}: compact enough to occupy a single context position, yet expressive enough to capture real behavioral structure. A behavior-aware vector-quantized variational autoencoder (VQ-VAE)~\citet{van2017neural} maps each buyer's de-identified historical clickstream to one of $K$ persona tokens, each encoding a coarse distinct behavioral profile. 
Because these tokens are learned end-to-end from real traffic, the distribution over token assignments naturally mirrors the true population mix of buyer types, enabling faithful reconstruction of store-level traffic patterns, something text-based methods fundamentally cannot provide. The learned tokens are added directly to the LLM vocabulary, making them compatible with token-based agents and assignable to new buyers with a single encoder forward pass. Because newly introduced persona tokens must be aligned with the pretrained model's existing semantic and action spaces, we introduce a two-stage persona-grounding procedure that decouples learning \emph{what} each token means from learning \emph{how} to act on it. This prevents shortcut learning from surface cues in the prompt and produces persona embeddings that transfer across unseen stores. Training uses only a small corpus disjoint from evaluation to encourage the model to learn reusable behavioral patterns rather than memorizing sessions. At inference, persona assignment scales to millions of buyers in seconds without retraining or per-store calibration. In summary, our contributions are:
\begin{enumerate}
    \item \textsc{SimPersona}, a framework that learns discrete persona tokens from raw clickstreams via a behavior-aware VQ-VAE, enabling scalable persona assignment, faithful reconstruction of buyer population distributions, and closed-loop agent simulation.
    \item  A two-stage persona-grounding framework that decouples \emph{what} each token means from \emph{how} to act on it, producing embeddings that generalize across storefronts without adaptation.
    \item An open-source data pipeline for converting raw clickstreams into buyer representations and agent-training traces, providing infrastructure for buyer behavior simulation.
    \item Evaluation on $8.37$M unseen buyers showing $78\%$ conversion alignment, statistically significant behavioral separation, and stronger instruction following than $8{\times}$ larger baseline.
\end{enumerate}

The remainder of the paper is organized as follows. \Cref{sec: method} details the proposed framework, \Cref{sec: results} evaluates it on $8.37$M buyers, and \Cref{sec:conclusion} discusses limitations and future directions.
\section{Proposed method}
\label{sec: method}
As detailed below, we build persona-conditioned shopping agents from raw clickstream data by developing a data pipeline for buyer representations and agent traces (\Cref{sec:data_pipeline}), a behavior-aware VQ-VAE that learns discrete personas and maps them to trainable tokens (\Cref{sec:vqvae}), and a multi-stage fine-tuning procedure that grounds these tokens in agent behavior (\Cref{sec:sft}).

\subsection{Data pipeline}
\label{sec:data_pipeline}

Existing buyer-simulation methods typically begin from curated inputs including benchmark tasks, explicit shopping preferences, or simulated user profiles rather than the raw clickstream logs available in production e-commerce systems~\citet{lu2025uxagent,wang2025opera,wang2025agentab}.
In real platforms, however, buyer behavior is observed through fragmented low-level events (page views, searches, cart mutations, and checkout actions) and these logs are optimized for analytics rather than for buyer modeling or agent training. As a result, they contain rich behavioral signal, but not in a form usable by representation-learning methods or LLM agents. 
Converting raw clickstreams into structured buyer representations and grounded multi-turn action traces is therefore a necessary but largely underexplored step.

We address this gap with a modular pipeline that performs a single enrichment pass over raw clickstreams and produces two complementary data types, \Cref{fig:data-pipeline}. It first produces compact aggregated buyer-level representations that combine behavioral statistics with semantic product context; enabling downstream tasks such as segmentation, clustering, and persona discovery. Then, it generates executable agent traces based on live storefront interactions. The traces are formatted as multi-turn sequences of DOM observations, structured actions, memory, and step-level reasoning for supervised fine tuning (SFT).  These components are detailed below in the context of persona learning (the pipeline itself is task-agnostic and open-source\footnote{URL will be available soon.})

\subsubsection{Buyer behavioral representation} \label{sec:raw_data}
We begin with raw platform logs containing page views, cart actions, search activities and other fine-grained interaction records. These records capture interactions but provide limited context about \emph{what} the buyer was interested in or how their actions relate across a session. Hence, we enrich the logs by augmenting them with the product catalog, collection directory, and search-query records; producing aggregated buyer--session data that combine behavioral signals with semantic context (e.g., product titles, collection names, and search queries), see \Cref{fig:enrichment}.

For persona discovery, we compress each buyer--session history into a single vector that captures both \emph{how} the buyer shops and \emph{what} they shop for.
To represent shopping style, we extract 16 scalar features organized into five groups: \emph{exposure \& volume} (total sessions, active days, session counts across product views, add-to-cart, checkout, search, and collection browsing), \emph{engagement} (total session duration and total product views), \emph{funnel} (add-to-cart, checkout, and browse-only rates), \emph{intent strength}, and \emph{dollar values} (average cart and order value); details provided in Appendix \ref{sec:app-data}. Product preference is captured by averaging the 768-dimensional product embeddings over each buyer's viewed, carted, and purchased items, then reducing each to 128 dimensions via PCA (retaining $\sim$85\% variance) so that the three semantic channels do not overwhelm the behavioral scalars. Three binary masks are included to show which embedding channels are observed. The resulting 403-dimensional vector provides a compact buyer representation for each buyer--shop pair (see \Cref{fig:vqvae_ex}).

Many downstream components require a shared notion of where a buyer lies in the conversion funnel. Instead of using task-specific heuristics, we define a single \emph{funnel stratification} from aggregated event counts. Each buyer is assigned, via a priority cascade extracted from their raw behavioral signals, to one of five strata: \textbf{A} purchasers, \textbf{B} checkout abandoners, \textbf{C} cart builders, \textbf{D} window shoppers, or \textbf{E} bouncers. We reuse this label in the pipeline as it guides intent derivation for simulation, supports stratified sampling during training, and provides diagnostics for representation quality (\Cref{sec:results-vqvae}).

\subsubsection{From clickstreams to agent traces.} \label{sec:traces}
Buyer representations capture behavioral variation but fail to inform an agent about how those behaviors sequentially unroll in a storefront. To address this issue, we replay real buyer sessions on live storefronts. From the enriched session tables generated in the previous step, we reconstruct each buyer's timestamped event sequence and use an LLM (GPT-5) to rewrite it as a natural-language navigation goal that preserves the structure of the original session. A separate agent (Gemini 3 Flash) then executes this goal on the live stores, interacting with real pages and  recording its trajectory in an SFT-ready format. Each generated training example consists of a shared \emph{system prompt} that defines the agent's role, output schema, and interaction rules, paired with a \emph{user prompt} containing the session's inferred intent, the buyer profile field (populated with the learned persona token), a cumulative progress log that serves as the agent's memory for reasoning and error recovery, and the current page's DOM snapshot. The corresponding \emph{assistant turn} contains a structured JSON action and a reasoning trace (\Cref{fig:sft-trace}). 
Each example thus teaches the model to interpret page state, track progress, recover from errors, and most importantly, navigate according to the learned buyer persona. An independent LLM judge (GPT-5) filters the corpus to retain only successful trajectories.

\subsection{Persona discovery via behavior-aware VQ-VAE}
\label{sec:vqvae}
Our goal is to use raw clickstreams to learn latent buyer behavior that is usable by a language model. To this end we build a VQ-VAE where the encoder compresses high-dimensional behavioral features and the quantization layer maps them to a finite codebook whose entries map one-to-one to new trainable tokens in an LLM's vocabulary.
Unlike distance-based clusterings such as $k$-means, our model is trained end-to-end with a behavior-aware contrastive objective and semantic auxiliary heads, encouraging codebook entries to reflect meaningful behavioral similarity rather than mere geometric proximity.
This discretization also enables \emph{distribution learning}, because the codebook is learned jointly over the full buyer population, the frequency with which buyers are assigned to each token provides a natural estimate of the latent distribution of buyer types. In Appendix~\ref{app:distribution}, we show that this learned distribution recovers aggregate population-level buyer behavior.

Encoder maps the input $\mathbf{x} \in \mathbb{R}^{d_{\text{in}}}$ through Linear$\to$ReLU$\to$Dropout blocks to a latent vector $\mathbf{z}_e \in \mathbb{R}^{D}$, which is quantized to its nearest codebook entry $\mathbf{e}_{k^*}$ from a learned codebook $\{\mathbf{e}_1, \dots, \mathbf{e}_K\} \subset \mathbb{R}^{D}$:
\begin{equation}
    k^* = \argmin_{k \in \{1,\dots,K\}} \|\mathbf{z}_e - \mathbf{e}_k\|_2^2,
    \qquad
    \mathbf{z}_q = \mathbf{e}_{k^*},
    \label{eq:vq}
\end{equation}
with gradients propagated via the straight-through estimator~\citet{van2017neural}. The decoder mirrors the encoder and reconstructs the input from $\mathbf{z}_q$. Codebook entries are initialized with $k$-means++~\citet{arthur2007kmeanspp} over encoder outputs and updated via exponential moving averages (EMA). To prevent codebook collapse~\citet{razavi2019generating}, we track an EMA of code usage and mark entries as dead when their count falls below a fraction $\alpha$ of the mean. After an initial warmup, dead entries are reinitialized from randomly sampled active encoder outputs. Details regarding hyperparameters are provided in Appendix \ref{sec:app-vqvae} (code is available in data pipeline repository).

Reconstruction loss does not guarantee semantically meaningful personas since similar features may correspond to buyers with different behaviors. We therefore augment the standard VQ-VAE objective with contrastive and auxiliary supervision to enforce semantic clustering:
\begin{equation}
    \mathcal{L} = \lambda_r\,\mathcal{L}_{\text{recon}}
                + \beta\,\mathcal{L}_{\text{commit}}
                + \lambda_c\,\mathcal{L}_{\text{contrastive}}
                + \lambda_a\bigl(\mathcal{L}_{\text{engage}}
                  + \mathcal{L}_{\text{explore}}
                  + \mathcal{L}_{\text{purchase}}\bigr).
    \label{eq:loss}
\end{equation}
Our reconstruction term ($\mathcal{L}_{\text{recon}}$) is group-aware. Scalar behavioral features are reconstructed with MSE, while product-preference embeddings are reconstructed with cosine distance to preserve semantic direction rather than magnitude. Also, to prevent the high-dimensional embedding channels from dominating the loss, we compute reconstruction at the level of semantic groups rather than raw dimensions, and we mask embedding terms when the corresponding interaction type is absent (a buyer who never added to cart incurs no loss on the carted-product embedding). The commitment loss
$\mathcal{L}_{\text{commit}} = \|\mathbf{z}_e - \mathrm{sg}(\mathbf{z}_q)\|_2^2$ keeps encoder outputs close to their assigned codes.

Reconstruction loss shapes what the codebook preserves, but provides no direct signal about \emph{which buyers should share a code}.  We propose a three-stage contrastive objective via InfoNCE~\citet{oord2018representation} that constructs positive pairs through progressively finer gates.  
\emph{Stage~1 (funnel gate)}: each buyer receives an ordinal signature from their highest interaction level (purchase$>$cart$>$view$>$none) where only same-signature buyers can form pairs.  
\emph{Stage~2 (product filter)}: within the same-signature pool, the top-$M$ peers are retained by cosine similarity in product-embedding space, ensuring paired buyers engage with similar products.  
\emph{Stage~3 (behavioral filter)}: from these candidate peers, the top-$F$ neighbors ($M>F$) are selected using Euclidean distance over exploration and engagement features (product views, search sessions, collection views, session duration), enforcing similar browsing style on top of similar funnel depth and product interest. All non-self samples remain in the InfoNCE denominator, so the model is simultaneously encouraged to pull behaviorally aligned buyers together and push apart incompatible ones. See Appendix \ref{sec:app-contrastive} for formulation.

To further encourage the codebook to capture interpretable shopping behavior, we attach three auxiliary classification heads to \(\mathbf{z}_q\). Each head is a cross-entropy loss to predict a coarse three-level label (low, medium, or high) along a behavioral axis: \emph{engagement depth}, \emph{exploration breadth}, and \emph{purchase intensity}. These auxiliary tasks provide supervised pressure for the codebook to preserve behavioral distinctions.
The labels are derived from buyer-level aggregate scores, with binning tailored to each target's distribution. Engagement is measured by total session duration and exploration by the number of search, collection-view, and product-view sessions; both are approximately log-normal, so we log-transform them and split into percentile-based terciles. Purchase intensity is defined as $8 \times \texttt{checkout\_sessions} + 3 \times \texttt{atc\_sessions}$ (see Appendix \ref{sec:app-weights}), which produces a discrete, heavily zero-inflated distribution with natural gaps between non-purchasers, light buyers, and heavy buyers. We place bin boundaries at these gaps to obtain three groups that align with distinct purchasing behaviors. To prevent the model from ignoring rare but important groups such as heavy purchasers, we apply inverse-frequency class weights $w_i = N/(B \cdot n_i)$, where $N$ is the training-set size, $B{=}3$, and $n_i$ is the count in bin $i$. After training, these heads also provide interpretable labels for each codebook entry, making it straightforward to inspect what each persona code represents (see Appendix \ref{sec:app-contrastive}).

\subsection{Multi-stage supervised fine-tuning}
\label{sec:sft}
The VQ-VAE produces discrete persona indices, but these indices have no inherent semantics for the language model. We therefore extend the Qwen3-14B-Base \citet{yang2025qwen3} tokenizer with $K$ special tokens (\texttt{<|persona\_0|>}, \dots, \texttt{<|persona\_\{K{-}1\}|>}), one for each codebook entry. This creates a central learning problem as the model must learn to \emph{condition its actions on the persona token} rather than on surface cues in the prompt.  End-to-end fine-tuning fails at this stage: randomly initialized
embeddings inject noise early in training, while the model quickly discovers that simple lexical cues in the inferred intent such as ``buy'' versus ``browse'' provide an easier shortcut, allowing it to ignore the persona token. To address this, we propose a two-stage training framework that explicitly separates learning \emph{what each persona token means} from learning \emph{how to act on it}. We ablate this choice in Appendix \ref{sec:app-staging} and find that two stage training is essential for robustness.

In Stage 1 (persona grounding), we freeze the pretrained backbone and train only the $K$ new embeddings. Each training example is a browsing trace from the data pipeline
(\Cref{sec:traces,fig:sft-stage1}) in which the buyer profile field contains the persona token, and the goal is an intent for session. Crucially, all goals in this stage are \emph{intent-neutral}: short
product-interest statements such as ``You are interested in skirts,'' with the category inferred from the buyer's interaction history.  Because the goal never reveals whether the buyer will browse, cart, or purchase, the only signal distinguishing a high-conversion token from a low-conversion token is the statistical pattern of actions across training traces. This signal forces the embedding to absorb the behavioral meaning of its persona. At the end of Stage 1 each persona embedding has converged to a distinct, stable region of the representation space without perturbing the pretrained weights.

In Stage~2 (action-oriented fine-tuning), we unfreeze the backbone and continue training on richer, goal-directed traces.  Goals now carry explicit intent derived from the buyer's funnel stratum: sessions from strata A--C receive prompts such as ``You are here to buy skirts,'' while stratum-D sessions receive ``You are here to browse skirts.''  
Since the intent is now present in both the goal \emph{and} the persona token, the model must learn to fuse two complementary signals: the token encodes the buyer's general behavioral profile while the goal specifies the intent of \emph{this} session (see \Cref{fig:sft-stage2}).  Neither signal is redundant, e.g., a high-purchase token paired with a browse goal should still produce exploratory behavior and so the model cannot collapse to either cue alone. Both stages draw from a small pool of shops fully disjoint from evaluation.  Because Stage~1 anchors the embeddings via aggregate behavioral co-occurrence against a frozen backbone, the learned tokens transfer to unseen storefronts without adaptation; we confirm this in \Cref{sec:results-heads} (architectural details in Appendix \ref{sec:app-sft-examples}).

\section{Discussion and results} \label{sec: results}
In this section, we evaluate our method along four axes: persona clustering quality (\Cref{sec:results-vqvae}), conversion alignment with real buyers (\Cref{sec:results-alignment}), fine-grained behavioral fidelity (\Cref{sec:results-heads}), and instruction-following performance against GPT-OSS-120B (\Cref{sec:results-instruction}).

\subsection{Does our clustering learn behaviorally meaningful personas?}
\label{sec:results-vqvae}
We train the VQ-VAE on a balanced subset of $39$ shops with sufficient support across strata A--E (\Cref{sec:data_pipeline}). For each shop, we sample up to $1{,}500$ buyer-shop pairs, capped at $300$ per stratum, to avoid over-representing either large shops or dominant buyer types. After removing stratum~E (bouncers), which contains limited behavioral signal, the resulting dataset contains $44{,}559$ buyer-shop pairs, split $85/15$ into training and validation sets. The model takes as input $16$ behavioral scalars together with three product embeddings explained in Appendix \ref{sec:app-data} and \Cref{sec:data_pipeline}. 

We evaluate the learned VQ-VAE codebook by comparing it against MiniBatch $k$-means trained on the same buyer representations with $K{=}256$ clusters. We assess four complementary aspects of cluster quality. \emph{Stratum purity} measures whether each cluster respects the funnel strata from \Cref{sec:data_pipeline}; we flag clusters with \emph{incompatible mixing}, where high-funnel buyers (strata A--C) and window shoppers (stratum~D) co-occur in substantial proportions. \emph{Auxiliary-head coherence} measures behavioral consistency along the three supervised axes from \Cref{sec:vqvae} by reporting the percentage of buyers assigned to codes that span non-adjacent bins (e.g., low and high). \emph{Pairwise cosine similarity} quantifies how alike buyers within the same cluster are in the original feature space. Finally, the \emph{Calinski--Harabasz index}~\citet{calinski1974dendrite} summarizes the ratio of between-cluster to within-cluster dispersion; higher values indicate tighter and better separated clusters (see \Cref{sec:ch}). 
\begin{table}[t]
  \caption{VQ-VAE vs.\ MiniBatch $k$-means ($K{=}256$). Coherence: $\%$ of buyers assigned to clusters spanning non-adjacent bins ($\downarrow$).}
  \label{tab:vqvae_comparison}
   \vspace{5pt}
  \centering
  \begin{tabular}{l cc ccc cc}
    \toprule
    & \multicolumn{2}{c}{\emph{Stratum purity}} & \multicolumn{3}{c}{\emph{Coherence (\% incoh.\ $\downarrow$)}} & \multicolumn{2}{c}{\emph{Separation}} \\
    \cmidrule(lr){2-3} \cmidrule(lr){4-6} \cmidrule(lr){7-8}
    Method
    & \makecell{Mean\\purity}
    & \makecell{Incompat.\\mixing}
    & Engage
    & Explore
    & Purchase
    & \makecell{PW\\cosine}
    & \makecell{Calinski--\\Harabasz} \\
    \midrule
    $k$-means & 78.9\% & 6 & 66.7\% & 4.6\% & 5.4\% & 0.722 & 173.8 \\
    VQ-VAE    & \textbf{84.5\%} & \textbf{0} & \textbf{0.5\%} & \textbf{0.0\%} & \textbf{0.0\%} & \textbf{0.774} & \textbf{206.5} \\
    \bottomrule
  \end{tabular}
\end{table}

Based on the results summarized in \Cref{tab:vqvae_comparison}, the most striking difference is in behavioral semantics.  Because $k$-means has no notion of what the features \emph{mean}, it simply minimizes Euclidean distance and freely merges buyers at opposite ends of a behavioral axis whenever doing so reduces geometric cost.  This is visible in the auxiliary-head coherence: $66.7\%$ of buyers fall in engagement-incoherent codes under $k$-means, meaning low-engagement and high-engagement buyers are routinely assigned to the same cluster.  VQ-VAE, by contrast, keeps incoherence below $0.5\%$ on all three heads.  The same pattern appears in stratum purity ($84.5\%$ vs.\ $78.9\%$) and incompatible mixing ($0$ vs.\ $6$ severely mixed codes): without semantic supervision, $k$-means conflates fundamentally different buyer types. On the combined coherence metrics (pairwise cosine similarity and CH index), VQ-VAE leads across the board, the modest margin is because $k$-means achieves strong tightness on the product-embedding dimensions, which dominate the concatenation by sheer count.  But this geometric tightness is hollow as it comes at the cost of the behavioral coherence that matters for downstream persona conditioning. 

\subsection{Do persona-conditioned agents match humans' conversion rates?}
\label{sec:results-alignment}
Fine-tuning uses the same $39$ training shops but only a small subset of approximately $3{,}600$ sessions. We deliberately downsample this set to ensure that every codebook entry is represented while keeping the corpus small enough to discourage memorization. 
At inference time, we evaluate on the full buyer population, spanning $8.37$M unique buyers across $42$ shops, none of which overlap with the training shops. This setup provides a direct test of whether the learned persona tokens generalize to unseen storefronts. Each agent is deployed on the \emph{same live} storefront visited by its real counterpart and is conditioned on two inputs: a persona token specifying the buyer profile, and a shopping intent (same as Stage 2 in \Cref{sec:sft}). The agent then navigates the storefront autonomously, and we ask whether its conversion behavior matches that of the real buyers represented by the same persona.

Answering this question requires a metric beyond standard task-completion benchmarks~\citet{zhou2024webarena,deng2023mind2web,yao2022webshop}, where success is binary. For buyer simulation, what matters is not whether the agent \emph{can} convert, but whether it converts at the \emph{same rate} as the real buyers it represents. We therefore define:
\begin{align*}
\text{ATC alignment} &= 1 - \left| \text{ATC}_{\text{real}} - \text{ATC}_{\text{agent}} \right|, \\
\text{Purchase alignment} &= 1 - \left| \text{PUR}_{\text{real}} - \text{PUR}_{\text{agent}} \right|, \\
\text{Action Rate Alignment (ARA)} &= 0.5\times \text{ATC alignment} + 0.5 \times \text{Purchase alignment},
\end{align*}
where $\text{ATC}$ and $\text{PUR}$ are the add-to-cart and checkout fractions per $(\text{shop}, \text{token})$ pair, stratified by funnel stratum to prevent dominant non-converting populations from washing out rare but important buyer types.
To verify that alignment is driven by the persona token itself and not by shop-level tendencies, we compare \emph{correct} pairings where an agent conditioned on token $a$ is evaluated against buyers assigned to token $a$, against two mismatch baselines: \emph{all-mismatch}, where the same agent is compared against buyers of every other token $b \neq a$ on that shop and averaged, and \emph{random-mismatch}, where the agent is compared against buyers of one randomly chosen incorrect token. If persona tokens carry meaningful signal, correct pairings should consistently outperform both baselines (more ablations and analysis are provided in Appendix \ref{sec:app-token-ablation}).

\begin{table}[t]
\caption{Conversion alignment by funnel stratum. Correct: matched token. All-Mis: averaged over all wrong tokens. 1-R: one random wrong token.}
\label{tab:alignment}
 \vspace{5pt}
\centering
\begin{tabular}{lccccccccc}
\toprule
& \multicolumn{3}{c}{ATC Alignment} &
\multicolumn{3}{c}{Purchase Alignment} &
\multicolumn{3}{c}{ARA} \\
\cmidrule(lr){2-4} \cmidrule(lr){5-7} \cmidrule(lr){8-10}
Stratum &
Correct & All-Mis& 1-R &
Correct & All-Mis & 1-R &
Correct & All-Mis & 1-R \\
\midrule
A &  .706 & .496 & .522 & .742 & .516 & .537 & .724 & .506 & .530 \\
B &  .736 & .465 & .429 & .760 & .598 & .544 & .748 & .532 & .487 \\
C &  .767 & .438 & .442 & .674 & .755 & .720 & .721 & .597 & .581 \\
D &  .942 & .653 & .644 & .965 & .792 & .769 & .954 & .723 & .707 \\
\midrule
Stratified & \textbf{.788} & .513 & .509 & \textbf{.785} & .665 & .643 & \textbf{.787} & .589 & .576 \\
\bottomrule
\end{tabular}%
\end{table}
\Cref{tab:alignment} breaks down alignment by funnel stratum.  Stratum~D is the easiest to match, since these buyers rarely add to cart or proceed to checkout; the target behavior is largely to browse without converting, and the agent reaches $95.1\%$ ARA. The task becomes harder deeper in the funnel. Stratum~C requires the agent to add to cart at the right rate \emph{without} proceeding to checkout, while strata~B and~A demand even finer control over when to cart, when to continue through checkout, and when to complete a purchase. These higher-funnel groups are also more behaviorally diverse, yet the agent still achieves strong alignment.
Critically, correct pairings outperform mismatch baselines by $21$--$24$pp across every stratum, with the widest gap on add-to-cart alignment; a discretionary, persona-specific action with high variance across buyer types. The stratified ARA of $78.7\%$ versus $57.4\%$ under mismatch confirms that a single discrete token meaningfully steers conversion behavior to match the real buyer population, not just at the easy end of the funnel but across all levels of behavioral complexity.

\subsection{Do persona-conditioned agents preserve behavioral dimensions?} \label{sec:results-heads}
In this section, we ask a finer-grained question: do the behavioral dimensions learned by the auxiliary heads also manifest in the agent's simulated behavior? As explained in \Cref{sec:vqvae}, each persona token encodes a Low/Medium/High bin on three behavioral axes. If these latent distinctions are preserved end-to-end, tokens in higher bins should produce systematically different simulated behavior than those in lower bins. We test this by grouping tokens according to their auxiliary-head bin and comparing simulated behavioral scores across groups using four complementary statistics: Cohen's $d$ for the standardized \textsc{Low}-vs-\textsc{High} effect size~\citet{cohen1988statistical}, Welch's $t$-test for pairwise mean differences~\citet{welch1947generalization}, the Kruskal--Wallis test for any difference across the three groups~\citet{kruskal1952use}, and a permutation test with $10{,}000$ random label reassignments~\citet{fisher1935design}.

Purchase intensity is the clearest example of why persona conditioning matters beyond coarse task intent. In our simulation setup, the intent specifies whether a session is broadly \emph{buy}- or \emph{browse}-oriented, but this single bit groups fundamentally different buyers across strata A, B, and C. Without persona conditioning, the agent has no way to distinguish these behaviors; it can only produce a single ``average buyer'' response to the buy prompt. The persona token resolves this ambiguity. As shown in the funnel progression section of \Cref{tab:behavioral-heads}, Low-purchase tokens almost never trigger conversion actions, Medium tokens convert at moderate rates, and High tokens convert routinely, all under the \emph{same} buy-oriented intent. This separation is not marginal: the mean purchase intensity score for High tokens is over $100{\times}$ that of Low tokens (\Cref{tab:behavioral-heads}), confirmed by a very large effect size (Cohen's $d = 2.80$) and highly significant statistical tests across all four measures ($p < 10^{-14}$). 
\begin{table}[t]
\centering
\caption{Behavioral separation across persona-token bins. Mean score is the average event count per simulation. Statistical significance is assessed at $\alpha=0.05$.}
\label{tab:behavioral-heads}
 \vspace{5pt}
\centering
\begin{tabular}{l ccc cccc}
\toprule
& \multicolumn{3}{c}{\emph{Mean score per bin}} & \multicolumn{4}{c}{\emph{Statistical tests}} \\
\cmidrule(lr){2-4} \cmidrule(lr){5-8}
\textbf{Head}
& Low & Med. & High
& Cohen's $d$ & $t$-test $p$ & KW $p$ & Perm.\ $p$ \\
\midrule
Purchase    & 0.06 & 2.94 & 7.25 & 2.80 & $5.5{\times}10^{-22}$ & $4.5{\times}10^{-14}$ & ${<}\,0.0001$ \\
Engagement  & 1.55 & 2.00 & 2.41 & 0.76 & $4.1{\times}10^{-3}$  & $1.2{\times}10^{-2}$  & $0.003$ \\
Exploration & 1.34 & 1.29 & 1.39 & 0.10 & $0.81$                & $0.78$                & $0.48$ \\
\midrule
\multicolumn{8}{l}{\emph{Purchase-bin funnel progression}} \\
\midrule
\textbf{Purchase bin} & \multicolumn{2}{c}{Low} & \multicolumn{2}{c}{Medium} & \multicolumn{2}{c}{High} & \\
\cmidrule(lr){2-3} \cmidrule(lr){4-5} \cmidrule(lr){6-7}
\textbf{Action} & \multicolumn{2}{c}{Rate} & \multicolumn{2}{c}{Rate} & \multicolumn{2}{c}{Rate} & \\
Add-to-Cart & \multicolumn{2}{c}{1.0\%} & \multicolumn{2}{c}{29.4\%} & \multicolumn{2}{c}{72.0\%} & \\
Checkout    & \multicolumn{2}{c}{0.3\%} & \multicolumn{2}{c}{13.7\%} & \multicolumn{2}{c}{43.3\%} & \\
\bottomrule
\end{tabular}
\end{table}

Engagement provides a second, distinct form of behavioral transfer. The VQ-VAE engagement head is defined from total session duration over a buyer's history, but simulated sessions lack wall-clock time. We therefore evaluate engagement using a behavioral proxy: the total number of actions the agent takes per session. 
As shown in \Cref{tab:behavioral-heads}, the mean engagement score rises monotonically from Low to High tokens, indicating that the persona token shapes not only \emph{whether} the agent converts, but also \emph{how actively} it participates in the session; High-engagement tokens produce agents that browse more pages and persist longer, while Low-engagement tokens produce agents that act quickly and leave. 
All four statistical tests confirm this separation is significant among all three levels (\Cref{tab:behavioral-heads}), with the permutation test ($p = 0.003$) providing the strongest evidence: fewer than $3$ in $1{,}000$ random label shuffles produced a gradient as strong as the observed one.

Exploration is the only auxiliary head that does not transfer reliably to agent behavior (\Cref{tab:behavioral-heads}). Mean scores are nearly flat across bins (${\sim}1.3$ events), with negligible effect size (Cohen's $d = 0.10$) and no significant differences. Several factors may contribute to this weak separation. First, the underlying human signal is itself subtle: even among real buyers, Medium and High exploration differ by only $0.12$ events per session, making the distinction difficult to recover from a single session. Second, the label distribution is heavily skewed, because our exploration metric aggregates product views, searches, and collection browsing over buyers who actively engaged (excluding bouncers), the VQ-VAE assigns most tokens to the High bin ($72\%$ High vs.\ $5\%$ Low), potentially leaving insufficient contrast to learn the distinction. Third, our simulation setup introduces a floor effect; every agent receives a goal naming a product category, so even Low-exploration personas must perform at least one search or product view to pursue their objective, eliminating near-zero sessions and pushing all bins toward a similar baseline of ${\sim}1.3$ events.

\subsection{Does fine-tuning improve instruction following?}
\begin{table}[t]
\caption{Instruction-following performance on $70$ deterministic tasks ($10$ repetitions each).}
\label{tab:instruction-following}
 \vspace{5pt}
\centering
\begin{tabular}{lccc}
\toprule
\textbf{Metric} & \textbf{\textsc{SimPersona}} & \textbf{GPT-OSS} & $\boldsymbol{\Delta}$ \\
\midrule
Avg steps per run          & 14.9   & 11.4   & +3.5 \\
Self-reported goal reached & 77.7\% & 93.9\% & $-$16.2pp \\
\midrule
\multicolumn{4}{l}{\emph{Funnel progression (required action reached)}} \\
\midrule
Cart tasks $\to$ added to cart        & 91.4\% & 81.7\% & +9.7pp \\
Checkout tasks $\to$ reached checkout & 76.9\% & 53.1\% & +23.8pp \\
Purchase tasks $\to$ reached checkout & 59.3\% & 52.2\% & +7.1pp \\
\bottomrule
\end{tabular}
\end{table}
\label{sec:results-instruction}
Beyond behavioral alignment, we ask whether SimPersona also improves the agent's ability to execute shopping tasks in live environments. We compare SimPersona against GPT-OSS-120B, a strong general-purpose baseline, on deterministic goal-oriented tasks executed on live storefronts. Rather than using a fixed benchmark, we derive each task from a sampled buyer's historical session so that task difficulty reflects the underlying behavioral profile: low-engagement buyers receive short lookup-style tasks, while more engaged buyers receive longer flows. We uniformly sample $70$ such tasks to cover persona-bin combinations, conversion outcomes, and complexity levels ranging from $3$ navigation actions to more than $20$ sequential steps (\Cref{tab:task-diversity}). Each task is repeated $10$ times per model, and an independent LLM judge (GPT-5) determines whether the objective was completed.

\Cref{tab:instruction-following} shows that \textsc{SimPersona} outperforms GPT-OSS on every action metric despite having nearly $8{\times}$ fewer parameters, with the advantage scaling with task difficulty. On simple cart tasks both models perform well, but the gap widens sharply on checkout tasks ($+23.8$pp), which require chaining product selection, cart management, and multi-page form navigation in a single session. Even on purchase tasks which are the hardest category, demanding end-to-end transactions within $30$ steps, \textsc{SimPersona} progresses further through the funnel. It also takes more steps per simulation ($14.9$ vs.\ $11.4$), reflecting persistence through complex navigation rather than early termination. The most revealing contrast is in \emph{how each model fails}: GPT-OSS declares \texttt{goal\_reached} in $93.9\%$ of simulations yet frequently stops before completing the required action, whereas \textsc{SimPersona} self-reports less often ($77.7\%$) but is far better calibrated; when it claims success, it has actually progressed further through the funnel.

\section{Conclusion}
\label{sec:conclusion}
We presented \textsc{SimPersona}, a framework that bridges user modeling and agent simulation by learning discrete persona tokens from raw clickstreams via a behavior-aware VQ-VAE and grounding them in an LLM through two-stage fine-tuning. On live storefronts unseen during training, \textsc{SimPersona} reproduces real conversion patterns, separates behavioral dimensions across personas, and outperforms a $8{\times}$ larger baseline; all with a single token per buyer and no per-store calibration.

Several directions remain open for future work; exploration behavior is less cleanly separated, suggesting richer signals are needed. DOM-based page representations omit visual cues that shape real decisions, motivating multimodal perception, and the current pipeline does not fully utilize the learned VQ-VAE embeddings after assignment; using them to initialize LLM token embeddings could reduce or eliminate the first grounding stage. Additionally, newly opened stores before clickstream history accumulates remain unsupported, as the framework requires buyer history.

\bibliographystyle{plainnat}
\bibliography{ref}

\newpage
\appendix

\section{Data Pipeline}
\label{sec:app-data}
\begin{figure}[h]
  \centering
  \includegraphics[width=0.9\textwidth, trim=0 80 0 0, clip]{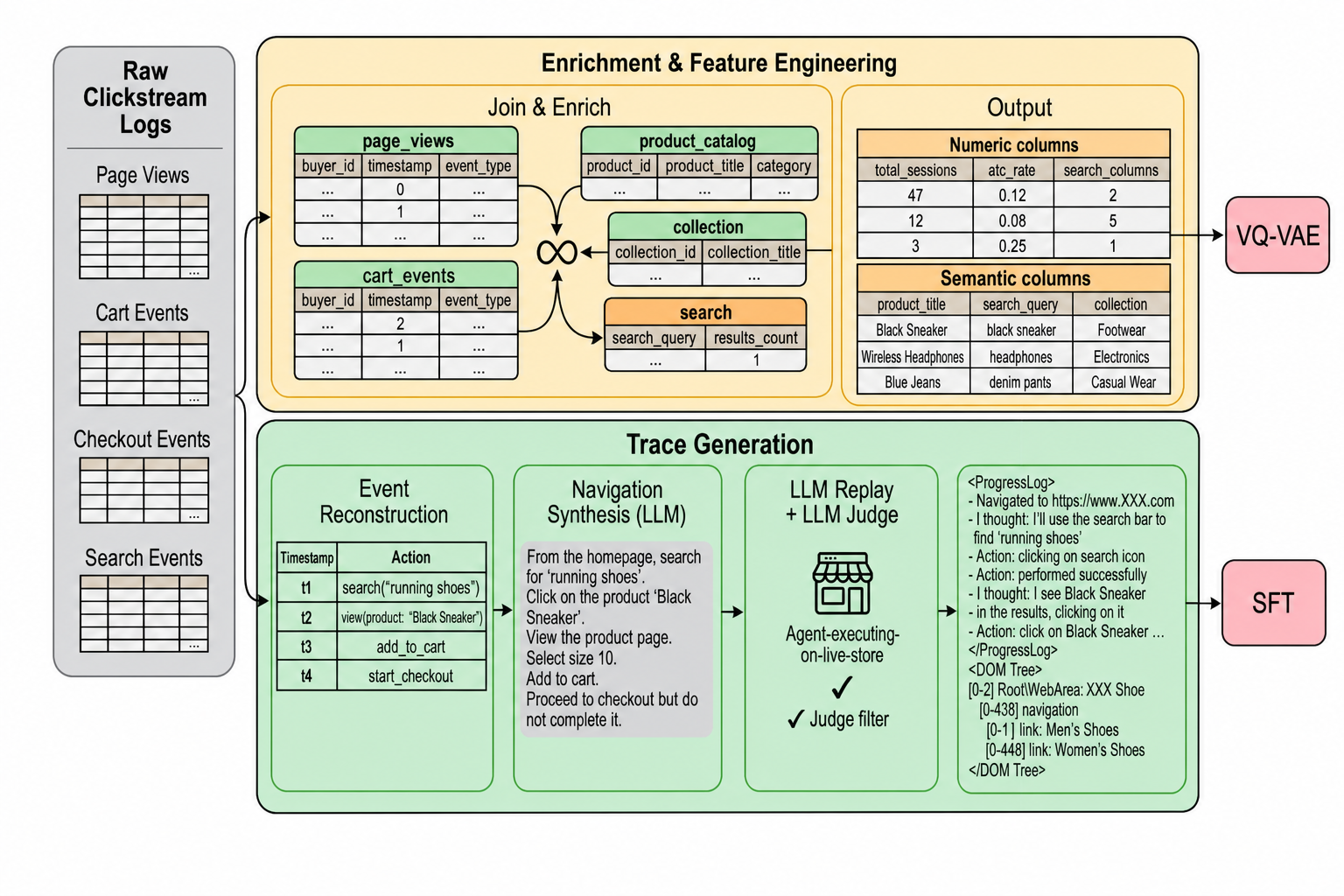}
  \caption{Data pipeline overview. A single enrichment pass over raw clickstream logs produces two outputs: (\textbf{top}) numeric and semantic buyer-level features for VQ-VAE persona discovery, and (\textbf{bottom}) executable multi-turn agent traces for SFT, generated by replaying real sessions on live storefronts.}
  \label{fig:data-pipeline}
\end{figure}

\begin{figure}[b]
  \centering
  \includegraphics[width=0.8\textwidth,trim=0 70 0 0, clip]{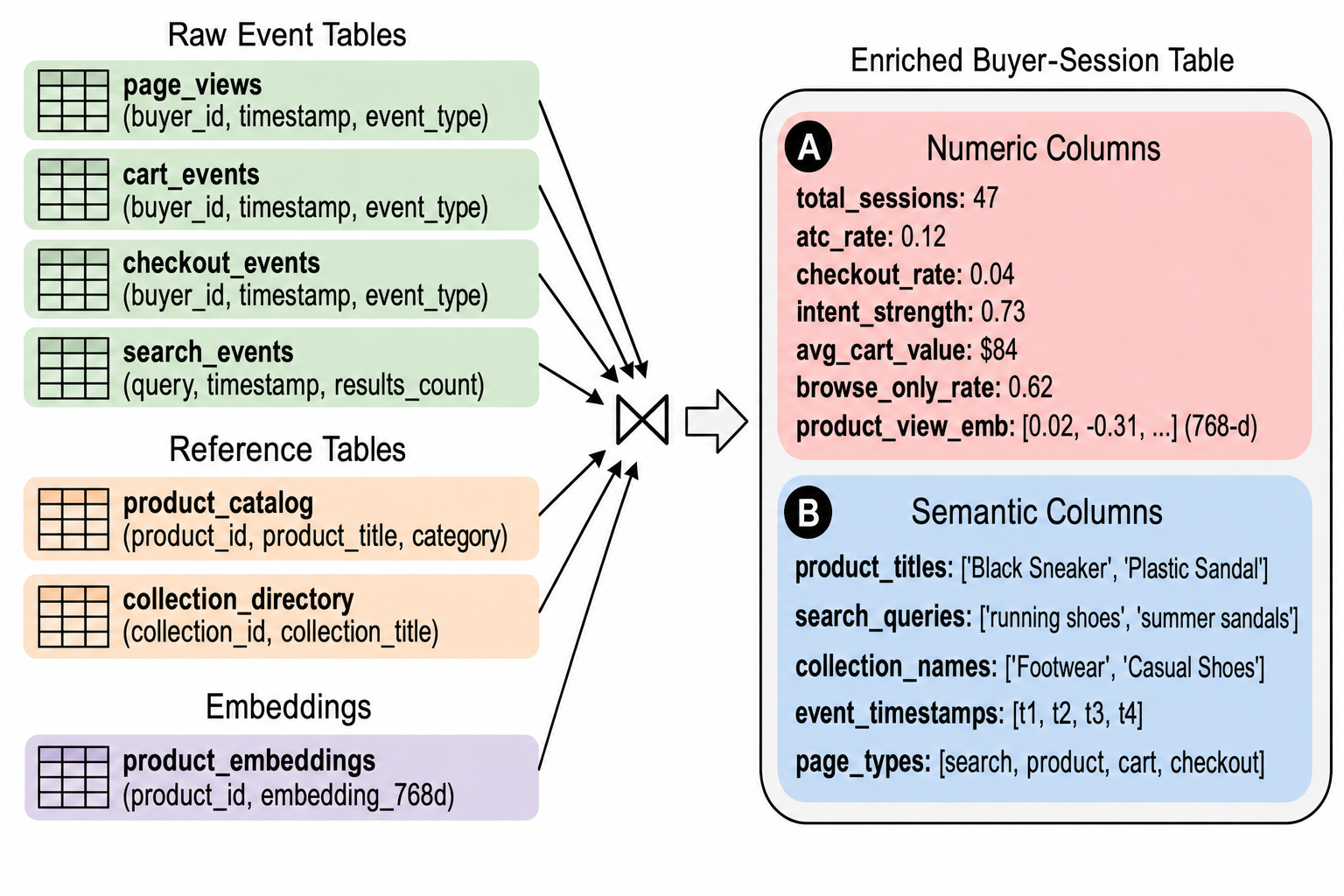}
  \caption{Data enrichment. Raw event-level tables are joined with the product catalog, collection directory, and product embeddings to produce an enriched buyer--session table containing (A) numeric behavioral features used for VQ-VAE persona discovery, and (B) semantic columns used for SFT trace generation.}
  \label{fig:enrichment}
\end{figure}

\Cref{fig:data-pipeline} illustrates our end-to-end data pipeline described in \Cref{sec:data_pipeline}. As mentioned in \Cref{sec:raw_data}, we join the de-identified raw inputs including the event-level information with the product catalog, collection directory, and search-query logs to produce enriched session-level records that combine numerical behavioral signals (e.g., session counts, conversion rates, session durations) with semantic information (e.g., product titles, search queries, collection names) shown in \Cref{fig:enrichment}.

From these enriched records, we extract the $16$ scalar behavioral features described in \Cref{sec:raw_data}. Among them, \emph{Intent strength} is a composite score calculated by weighting actions according to their commercial commitment:

\begin{equation}
    s_{\text{intent}} = \frac{w_{\text{atc}} \cdot n_{\text{atc}}
    + w_{\text{co}} \cdot n_{\text{co\_start}}
    + w_{\text{pur}} \cdot n_{\text{purchase}}}
    {n_{\text{sessions}}
    + w_{\text{atc}} \cdot n_{\text{atc}}
    + w_{\text{co}} \cdot n_{\text{co\_start}}
    + w_{\text{pur}} \cdot n_{\text{purchase}}}
    \label{eq:intent}
\end{equation}

where $n_{\text{atc}}$, $n_{\text{co\_start}}$, and $n_{\text{purchase}}$ count sessions containing an add-to-cart, checkout initiation, and completed purchase,
respectively. The weights $w_{\text{atc}}{=}3$, $w_{\text{co}}{=}5$, $w_{\text{pur}}{=}8$ are monotonically increasing with funnel depth, following the same coprime design principle used for the auxiliary purchase head (Appendix \ref{sec:app-weights}): $w_{\text{atc}}$ and $w_{\text{pur}}$ match the purchase composite weights, while $w_{\text{co}}{=}5$ is their midpoint, capturing checkout initiation as an intermediate commitment signal between add-to-cart and completed purchase.  The denominator ensures $s_{\text{intent}} \in [0, 1)$: a pure window shopper scores $0$, while a buyer who converts every session approaches ${\sim}0.94$.
\begin{table}[t]
\small
\caption{Feature normalization pipeline for the 403-dimensional VQ-VAE inputs.}
\label{tab:normalization}
 \vspace{5pt}
\centering
\begin{tabular}{@{}lrcl@{}}
\toprule
\textbf{Feature Group} & \textbf{Dims} & \textbf{Transform} & \textbf{Rationale} \\
\midrule
Exposure \& Volume & 8
  & $\log(1{+}x) \to \text{rob-}z$
  & Right-skewed counts \\
Engagement Depth & 2
  & $\log(1{+}x) \to \text{rob-}z$
  & Cumulative totals with heavy tails \\
Funnel Rates & 3
  & $\frac{n+\alpha}{N+\alpha+\beta} \to \mathrm{logit} \to \text{rob-}z$
  & Smoothing stabilizes small $N$ \\
Intent Strength & 1
  & $\mathrm{logit}_\epsilon \to \text{rob-}z$
  & Bounded composite score\\
Dollar Values & 2
  & $\log(1{+}x) \to \text{rob-}z$
  & \$0 for non-buyers, heavy tail \\
Product Embeddings & $3{\times}128$
  & dim-$z \to$ PCA\,$768{\to}128$
  & Heterogeneous dims; retains 85\% var \\
Embedding Masks & 3
  & binary (as-is)
  & Flags observed vs.\ zero-filled channels \\
\midrule
\textbf{Total} & \textbf{403} & & \\
\bottomrule
\end{tabular}
\end{table}
\begin{figure}[b!]
  \centering
  \includegraphics[width=0.8\textwidth, trim=0 80 0 0, clip]{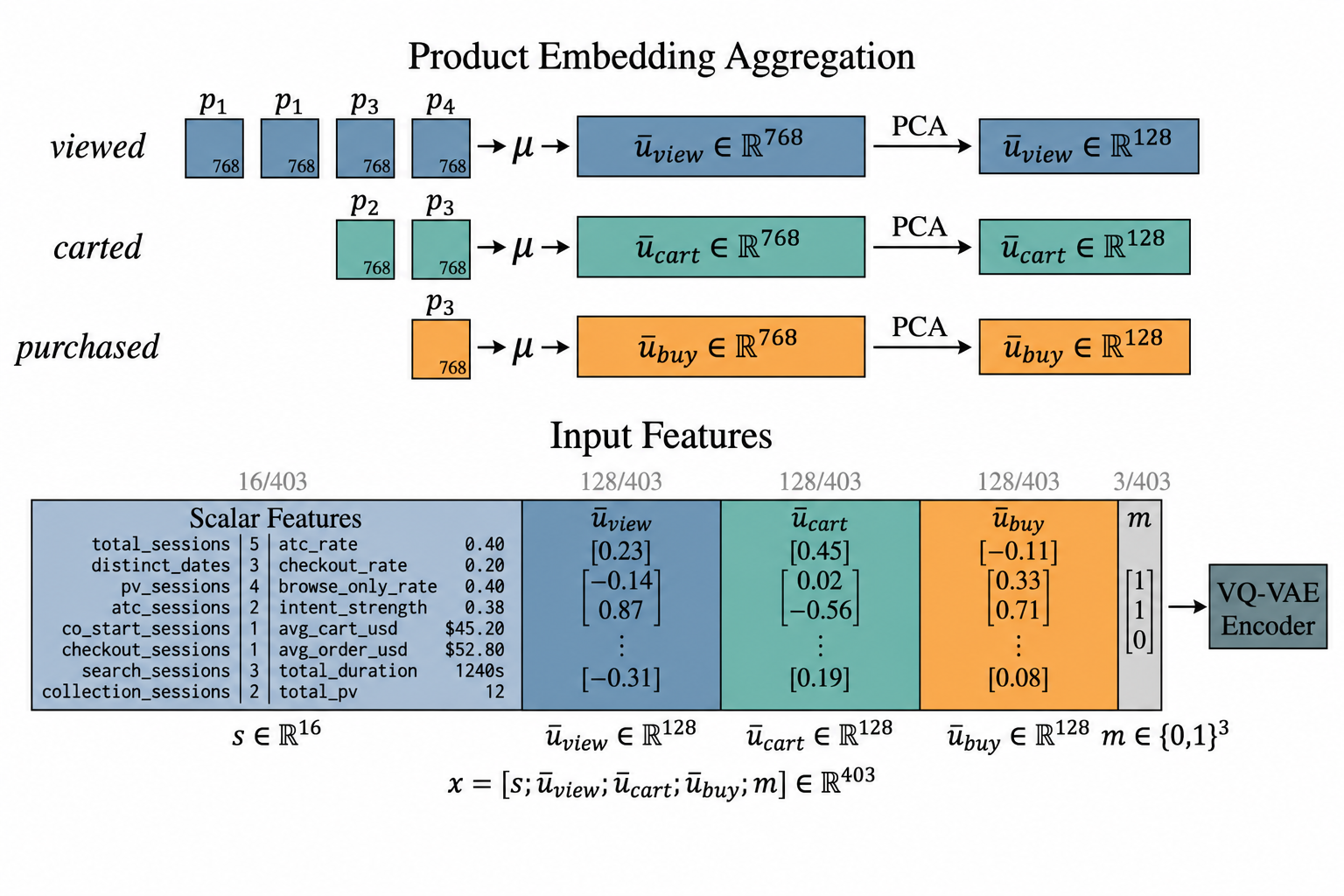}
  \caption{VQ-VAE input construction for a single buyer--shop pair. \textbf{top:} product embeddings are computed by averaging the $768$-dimensional catalog vectors of all products a buyer viewed, carted, or purchased, then compressed to $128$ dimensions via PCA. \textbf{Bottom:} the final $403$-dimensional input concatenates $16$ $z$-scored behavioral scalars, three $128$-d product embeddings, and a $3$-bit evidence mask indicating which product channels are present.}
  \label{fig:vqvae_ex}
\end{figure}
Because the $16$ features span different scales and distributions, each group is normalized with a tailored transform (\Cref{tab:normalization}). Heavy-tailed counts and dollar values are log-transformed and then robust $z$-scored, computed as $(x - \tilde{x})/\mathrm{IQR}$ with MAD$\,{\times}\,1.4826$ fallback when $\mathrm{IQR}{=}0$. Bounded rates are first Bayesian-smoothed to replace the raw rate $n/N$ with $\tilde{p} = (n{+}\alpha)/(N{+}\alpha{+}\beta)$ under a Laplace prior ($\alpha{=}\beta{=}1$) to prevent extreme logit values from buyers with few sessions, then logit-transformed and robust $z$-scored. These $16$ scalars are concatenated with three PCA-compressed product embeddings ($3 \times 128$ dimensions) and three binary evidence masks to form the $403$-dimensional VQ-VAE input illustrated in \Cref{fig:vqvae_ex}.

\begin{figure}[t]
  \centering
  \includegraphics[width=0.8\textwidth]{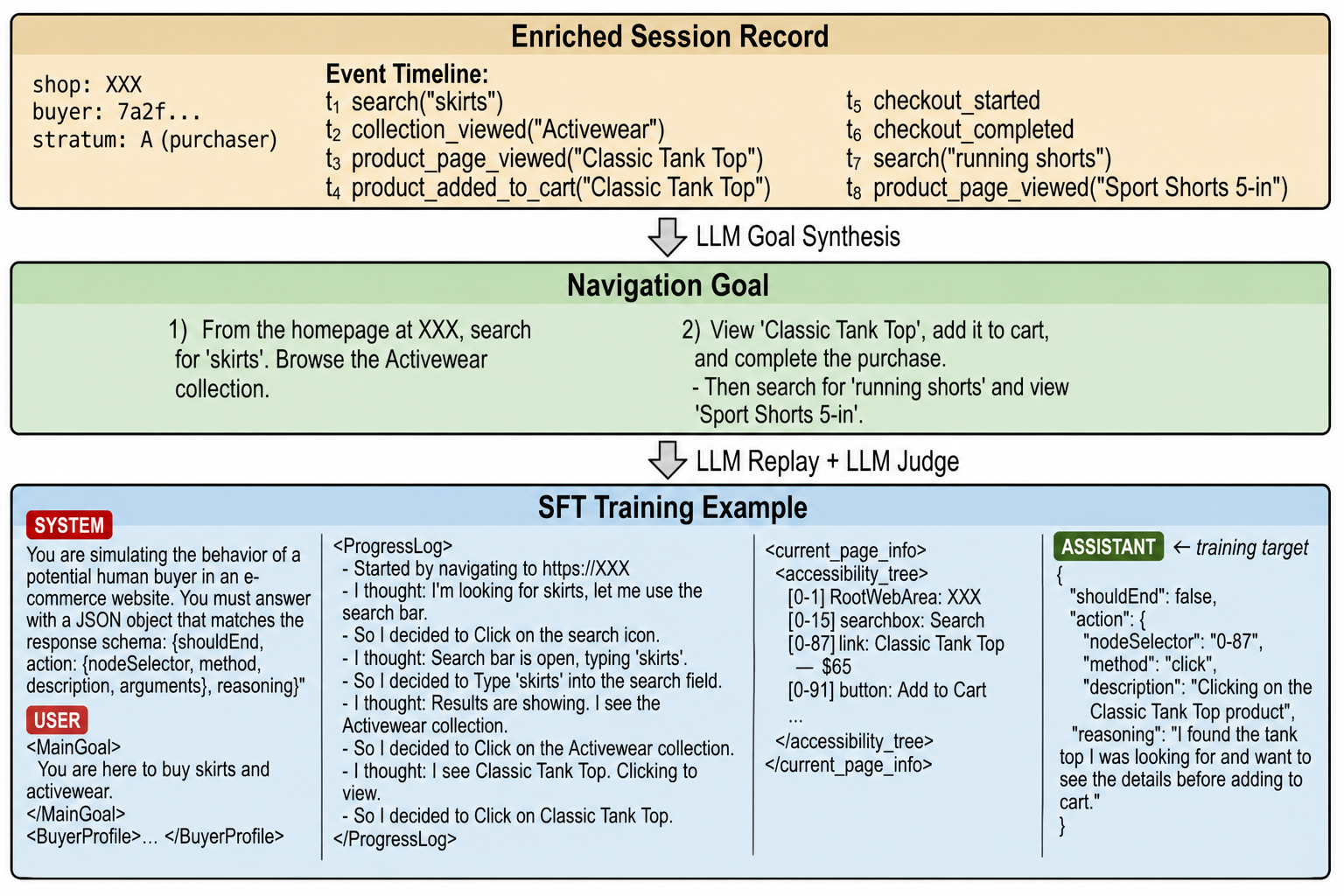}
  \caption{SFT trace generation from enriched clickstreams. \textbf{Row~1}: the enriched session record contains the buyer's event timeline with semantic labels (product names, collection titles, search queries) alongside metadata (shop ID, funnel stratum, persona token). \textbf{Row~2}: an LLM synthesizes the event sequence into a natural-language navigation goal, which is replayed on the live storefront by an agent and verified by an LLM judge. \textbf{Row~3}: the verified replay is converted into a multi-turn SFT example; system prompt defines the task format, user turn provides the navigation goal and DOM snapshot, and the assistant turn serves as the training target.}
  \label{fig:sft-trace}
\end{figure}

\Cref{fig:sft-trace} illustrates the trace generation pipeline described in \Cref{sec:traces}: a buyer's raw clickstream is first synthesized into a natural-language navigation goal, then an agent replays that goal on the live storefront, producing the multi-turn interaction traces used for SFT.

\section{Population Distribution Recovery}
\label{app:distribution}
A distinctive property of our proposed method is that it recovers not only \emph{which} behavioral types exist in the buyer population, but \emph{how they are distributed} within each store, a capability that is structurally absent from prompt-based persona methods, where persona types are defined by manual specification and no mechanism exists to estimate the mixing proportions ($p_s(k)$) from observed traffic.
In our method however, because each buyer $b$ is mapped to a discrete token via a single encoder forward pass followed by nearest-entry quantization $q(\cdot)$, the empirical population distribution over persona tokens for store $s$ is given by:
\begin{equation}
  \hat{p}_s(k) \;=\; \frac{1}{|\mathcal{B}_s|}
    \sum_{b \in \mathcal{B}_s} \mathbf{1}[\,q(z_b) = k\,],
    \qquad k = 1,\dots,K,
\end{equation}
where $\mathcal{B}_s$ is the set of buyers observed at that store. This distribution is the sampling measure used to construct simulated buyer populations: agents are assigned persona tokens in proportion to $\hat{p}_s(k)$. As a result, store-level simulation statistics such as add-to-cart rate, checkout rate, and navigation depth depend not only on how each token behaves, but also on whether the learned token mixture matches the real buyer population. If high-intent tokens are overrepresented, simulated conversion will be inflated; if low-intent tokens dominate incorrectly, it will be suppressed. Recovering $\hat{p}_s$ is therefore a necessary condition for faithful population-level simulation.

We test this property on the same $42$ held-out storefronts covering $8.37$M buyers used in \Cref{sec: results}, using two complementary analyses. The first asks whether the learned token mixture can recover each store's funnel-stage composition. The second asks whether the same token mixture can reconstruct continuous store-level behavioral features. Together, these tests evaluate whether the codebook captures real population structure rather than merely clustering individual sessions.
\begin{figure}[t]
  \centering
  \includegraphics[width=0.95\linewidth]{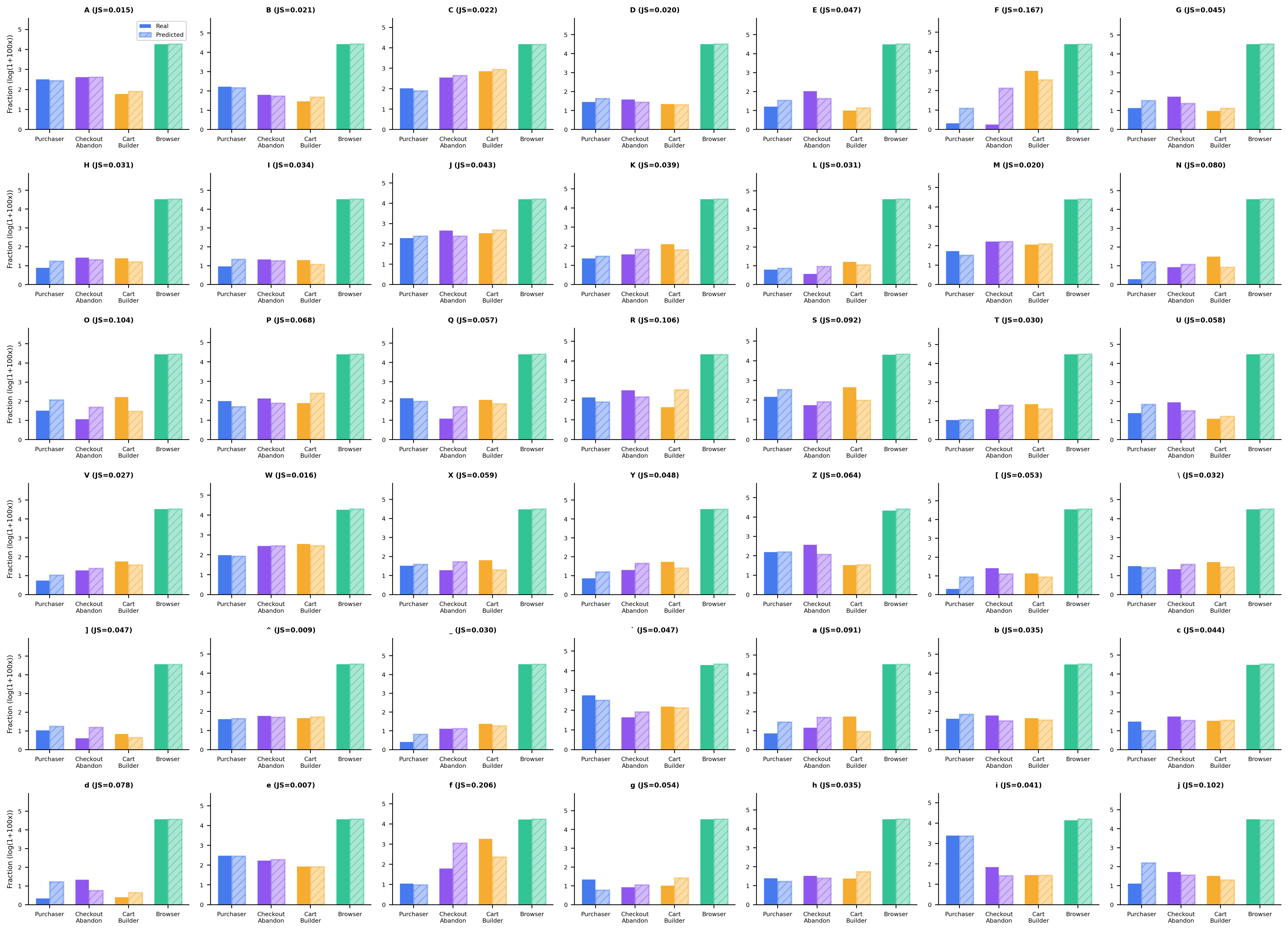}
  \caption{Stratum distribution recovery across all $42$ storefronts. Solid bars show the real funnel-stage composition and hatched bars show the distribution predicted from each store's VQ-VAE token distribution. Bar heights are shown as $\log(1+100x)$, where $x$ is the stratum fraction, to make low-frequency strata visible alongside the dominant browser segment. The predicted and real distributions remain closely matched under this visualization (mean JS divergence $=0.054$), confirming that the learned codebook faithfully recovers the buyer population mix.}
  \label{fig:stratum-recovery}
\end{figure}
We first evaluate recovery of funnel strata. As explained in \Cref{sec:raw_data}, each buyer is independently assigned to one of four rule-based strata from its historical behavior: Purchaser (A), Checkout Abandoner (B), Cart Builder (C), or Window Shopper (D). These labels are not used to train the VQ-VAE, which only sees continuous behavioral features, product embeddings, and evidence masks. This makes stratum recovery a stringent test: if the token distribution can reconstruct the stratum mix of a store, then the unsupervised codebook has recovered behaviorally meaningful population structure.

For each token $k$, we estimate its global stratum profile $P(\text{stratum}\mid k)$ from all buyers assigned to that token. We then predict the stratum distribution of store $s$ by mixing these token profiles according to the store's token distribution:
\begin{equation}
  \hat{P}_s(\text{stratum})
  =
  \sum_{k=1}^{K}
  \hat{p}_s(k)\,P(\text{stratum}\mid k).
\end{equation}
Intuitively, this asks if we know which persona tokens a store's buyers fall into, and we know what funnel stage each token typically represents, can we recover the store's real funnel-stage mix?
We measure the agreement using the Jensen--Shannon (JS) divergence~\citet{lin1991divergence}, a symmetric, bounded measure of similarity between two probability distributions.
JS divergence ranges from $0$ (identical distributions) to $1$ (completely disjoint support); values below $0.05$ indicate near-perfect agreement, while values above $0.15$ suggest meaningful distributional differences.

\Cref{fig:stratum-recovery} compares the predicted and real stratum distributions across all evaluation storefronts, from the largest store with $3.1$M buyers to the smallest with under $1{,}000$.
The mean Jensen--Shannon divergence is $0.054$ (median $0.045$), with $36$ of $42$ stores below $0.10$.
The recovery is equally accurate on stores with over one million buyers, where even small distributional errors would affect hundreds of thousands of sampled agents, and on stores with fewer than $10{,}000$ buyers, where the token distribution is estimated from a smaller sample.
Visually, the solid (real) and hatched (predicted) bars are nearly indistinguishable across stores, despite wide variation in buyer composition; from browser-dominated stores ($>90\%$ stratum~D) to stores with $30\%+$ purchasers.
This level of agreement between a hard rule-based labeling and an unsupervised discrete codebook trained on continuous features, confirms that the VQ-VAE codebook has learned behaviorally meaningful clusters whose population-level statistics faithfully reflect the real buyer distribution at scale.

\begin{figure}[t]
  \centering
  \includegraphics[width=0.8\linewidth]{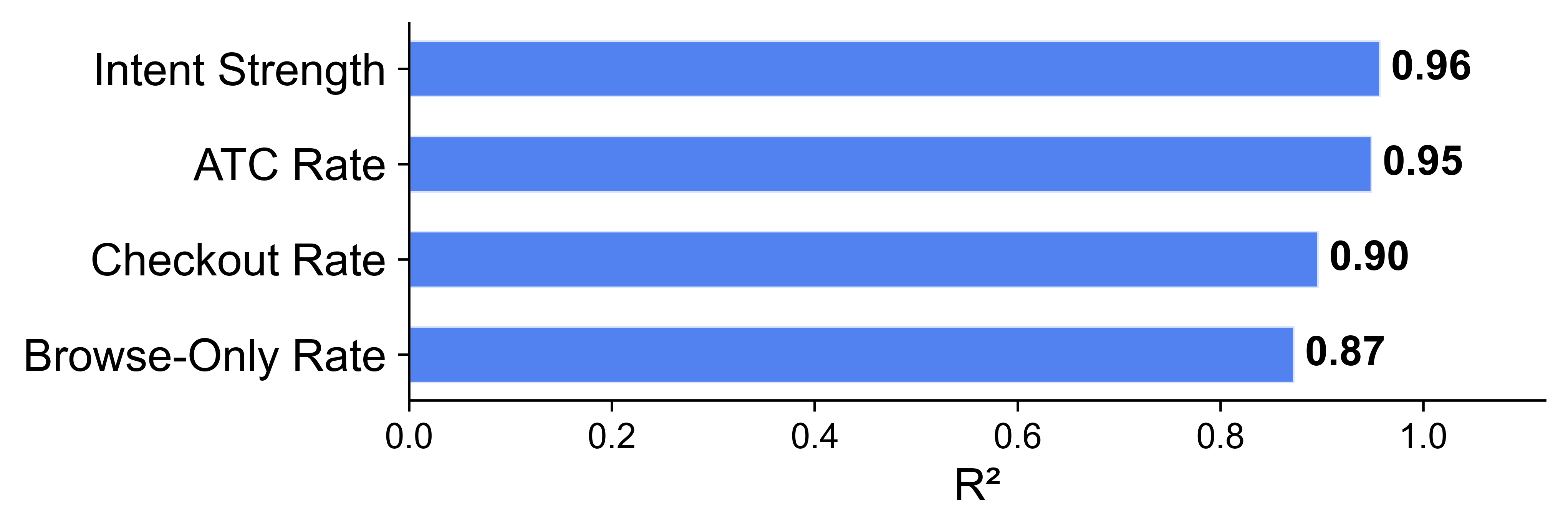}
  \caption{Store-level behavioral reconstruction from persona token distributions. The codebook achieves $R^2 \geq 0.87$ on the four features most critical to shopping simulation, confirming that it preserves behaviorally relevant population structure while remaining invariant to volume metrics.}
  \label{fig:feature-r2}
\end{figure}

We next test whether the same token distribution preserves continuous behavioral information beyond categorical strata.
For each token $k$, we compute its behavioral centroid $\bar{\mathbf{x}}_k \in \mathbb{R}^d$; the mean raw feature vector of all buyers assigned to that token across all stores.
For a given store $s$, we reconstruct its aggregate buyer profile by weighting these centroids by the store's token distribution:
\begin{equation}
  \hat{\mathbf{x}}_s =
  \sum_{k=1}^{K}
  \hat{p}_s(k)\,\bar{\mathbf{x}}_k .
\end{equation}
We compare $\hat{\mathbf{x}}_s$ to the true store-level mean $\mathbf{x}_s$ and report per-feature $R^2$ across all $42$ stores (\Cref{fig:feature-r2}).
$R^2$ measures the fraction of store-to-store variance in each feature that is explained by the token-weighted prediction: a value of $0.96$ means that $96\%$ of the cross-store variation in that feature can be recovered from the token distribution alone, without access to any individual buyer data.
The reconstruction is strongest on the behavioral dimensions most critical to simulation fidelity: intent strength ($R^2 = 0.96$), add-to-cart rate ($0.95$), checkout rate ($0.90$), and browse-only rate ($0.87$).
These are precisely the features that determine whether a simulated agent browses passively, expresses purchase interest, or advances toward checkout, the core axes along which persona tokens must differentiate buyer behavior.
Their consistently high $R^2$ ($\geq 0.87$) confirms that by knowing a store's token mix we can predict its real aggregate behavioral profile: if two stores differ substantially in how purchase-oriented their visitors are, the VQ-VAE token distribution captures that difference.

\section{VQ-VAE architecture and training details.}
\label{sec:app-vqvae}

As explained in \Cref{sec:vqvae}, codebook entries are initialized with $k$-means++~\citet{arthur2007kmeanspp} over encoder outputs from a full pass through the training set. During training, entries are updated via exponential moving averages rather than gradient descent:
\begin{equation}
    \mathbf{e}_k \leftarrow \gamma\,\mathbf{e}_k + (1-\gamma)\,\bar{\mathbf{z}}_k,
    \label{eq:ema}
\end{equation}
where $\bar{\mathbf{z}}_k$ is the mean of encoder outputs assigned to entry $k$ in the current batch and $\gamma \in [0,1)$ controls the memory of past assignments. To prevent codebook collapse~\citet{razavi2019generating}, we maintain an EMA of per-entry assignment counts $\hat{n}_k$ and declare an entry dead when
\begin{equation}
    \hat{n}_k \;<\; \alpha \cdot \frac{1}{K}\sum_{j=1}^{K}\hat{n}_j,
    \label{eq:dead-code}
\end{equation}
i.e., when its usage falls below a fraction $\alpha$ of the mean. Dead entries are reinitialized from randomly sampled encoder outputs plus small Gaussian noise. This reset activates only after a warmup period so that usage statistics reflect a stable latent space.

The encoder, codebook, and decoder have layer sizes $[403 \!\to\! 256 \!\to\! 128 \!\to\! 96]$, $K\!=\!256$ entries in $\mathbb{R}^{96}$, and a mirrored decoder, respectively. Reconstruction is down-weighted ($\lambda_r\!=\!0.3$) to prioritize behavioral partitioning over per-feature fidelity; remaining hyperparameters are $\beta\!=\!0.75$, $\lambda_c\!=\!0.15$, $\lambda_a\!=\!0.5$, EMA decay $\gamma\!=\!0.9$ (\Cref{eq:ema}), and dead-code replacement with threshold $\alpha\!=\!0.1$ every $50$ steps after a $100$-step warmup (\Cref{eq:dead-code}). All values were selected through hyperparameter search optimizing for behavioral purity of the resulting codebook entries. We highlight that the training data is constructed by aggregating user behavior over last $4$ months.

\section{Contrastive and auxiliary loss}
\label{sec:app-contrastive}

The contrastive term $\mathcal{L}_{\text{contrast}}$ is an InfoNCE loss~\citet{oord2018representation} computed over the encoder outputs $\mathbf{z}_e$:
\begin{equation}
    \mathcal{L}_{\text{contrast}} = -\frac{1}{|\mathcal{B}|}\sum_{i \in \mathcal{B}} \log \frac{\exp(\text{sim}(\mathbf{z}_e^{(i)}, \mathbf{z}_e^{(i+)}) / \tau)}{\sum_{j \in \mathcal{B}} \exp(\text{sim}(\mathbf{z}_e^{(i)}, \mathbf{z}_e^{(j)}) / \tau)},
    \label{eq:infonce}
\end{equation}
where $\text{sim}(\cdot,\cdot)$ denotes cosine similarity, $\tau$ is a temperature parameter, and $i+$ is the positive pair for anchor $i$. As explained in \Cref{sec:vqvae}, we define positive pairs through $3$ steps, and all non-self samples in the batch remain in the InfoNCE denominator (\Cref{eq:infonce}), so the model simultaneously pulls behaviorally aligned buyers together and pushes apart incompatible ones.

We set $\tau = 0.1$, $M = 10$, and $F = 3$, chosen via hyperparameter search over codebook purity optimization. The constraint $M > F$ is by design: the product filter retrieves a broad pool of peers who engage with similar merchandise, and the behavioral filter then narrows this pool to the $F$ whose browsing \emph{style} most closely matches the anchor buyer. If $F \geq M$, the behavioral stage would pass through every product-filtered candidate, collapsing the two-stage selection into a single filter and losing the ability to separate \emph{what} buyers shop for from \emph{how} they shop.

The auxiliary loss $\mathcal{L}_{\text{aux}}$ adds supervised pressure for the codebook to preserve behaviorally meaningful distinctions. Three single-layer linear heads operate on the quantized representation $\mathbf{z}_q$, each predicting a coarse three-way label (Low / Medium / High) for one behavioral axis:
\begin{equation}
\mathcal{L}_{\text{aux}}
= \sum_{h \in \{\text{engage},\, \text{explore},\, \text{purchase}\}} \mathcal{L}_h,
\label{eq:aux}
\end{equation}
where each $\mathcal{L}_h$ is a weighted cross-entropy loss:
\begin{equation}
\mathcal{L}_h
= -\sum_{c=1}^{3} w_c^{(h)}\, y_c^{(h)} \log \hat{p}_c^{(h)}(\mathbf{z}_q),
\qquad
\hat{p}^{(h)}(\mathbf{z}_q) = \mathrm{softmax}\!\bigl(\mathrm{MLP}_h(\mathbf{z}_q)\bigr).
\end{equation}
Here $y_c^{(h)} \in \{0,1\}$ is the one-hot ground-truth bin for head $h$, and $\hat{p}_c^{(h)}$ is the predicted probability for bin $c$. Because some bins are much smaller than others (e.g., heavy purchasers are rare), we apply inverse-frequency class weights $w_c^{(h)} = N / (3 \cdot n_c^{(h)})$, where $N$ is the training-set size and $n_c^{(h)}$ is the count in bin $c$. This ensures the loss does not ignore rare but behaviorally important groups.


\section{Auxiliary head weight selection}
\label{sec:app-weights}
\begin{table}[b]
\caption{Purchase-intensity bins induced by $s_{\mathrm{pur}} = 8\,n_{\texttt{checkout}} + 3\,n_{\texttt{atc}}$. The same partition is obtained for all $105$ tested monotone weight pairs satisfying $w_{\texttt{co}} > w_{\texttt{atc}} > 0$.}
\label{tab:purchase-bins}
 \vspace{5pt}
\centering
\begin{tabular}{llrccc}
\toprule
\textbf{Bin} & \textbf{Semantics} & $N$ & Checkout Rate & ATC Rate & Purchase Intensity \\
\midrule
0 & Browse only        & 17{,}453 & .000 & .000 & .000 \\
1 & Interest only      & 16{,}472 & .000 & .892 & .706 \\
2 & Purchase-committed & 10{,}634 & .533 & .536 & .776 \\
\bottomrule
\end{tabular}
\end{table}

The auxiliary purchase head (\Cref{sec:vqvae,sec:app-contrastive}) supervises the codebook to preserve coarse purchase-intensity structure. Each buyer receives a three-level label (\textit{Low / Medium / High}) derived from the composite score $s_{\mathrm{pur}} = w_{\texttt{co}}\, n_{\texttt{checkout}} + w_{\texttt{atc}}\, n_{\texttt{atc}}$, where $n_{\texttt{checkout}}$ and $n_{\texttt{atc}}$ count sessions containing a completed checkout and an add-to-cart, respectively. The only structural assumption is the funnel ordering $w_{\texttt{co}} > w_{\texttt{atc}} > 0$: checkout reflects stronger commitment than add-to-cart. The resulting integer scores are discretized into three bins by a count-based procedure that greedily merges adjacent values until three groups remain.

A key question is whether the auxiliary labels depend sensitively on the
particular numerical choice of $(w_{\texttt{co}}, w_{\texttt{atc}})$. To test
this, we swept all monotone integer pairs with
$w_{\texttt{co}} \in [2,15]$ and
$w_{\texttt{atc}} \in [1, w_{\texttt{co}}-1]$, for a total of $105$ candidate
configurations. On the $44{,}559$ VQ-VAE training buyers, every one of these
configurations produced exactly the same three-way partition
($17{,}453 / 16{,}472 / 10{,}634$ buyers). In other words, the induced labels are
not driven by any special tuning of the weights. Instead, the buyers already
separate naturally into three coarse behavioral groups: those with no commercial
activity, those with add-to-cart activity but no checkout behavior, and those
with checkout-bearing sessions. As long as the monotone ordering
$w_{\texttt{co}} > w_{\texttt{atc}}$ is respected, the exact ratio does not
change this partition.

Since the partition is invariant, $(w_{\texttt{co}}, w_{\texttt{atc}}) = (8, 3)$ is chosen for interpretability rather than as a tuned parameter. The values are coprime ($\gcd(8,3)=1$), which maximizes the number of composite scores that uniquely decompose into their constituent events: a score of $11$ can only mean one checkout plus one add-to-cart ($8 + 3$), and $6$ can only mean two add-to-carts ($3 \times 2$). A non-coprime pair like $(8, 2)$ would make scores like $16$ ambiguous; two checkouts ($8 \times 2$) or eight add-to-carts ($2 \times 8$), conflating fundamentally different funnel behaviors. Beyond coprimality, the gap between $8$ and $3$ is large enough that the smallest checkout-containing score ($8$) strictly exceeds two add-to-carts ($6$), cleanly separating repeated interest from genuine funnel progression in the low-count regime that dominates real buyer data. The resulting half-integer bin edges $[1.5, 4.5]$ further guarantee that no buyer's integer-valued score falls on a boundary, making every bin assignment unambiguous. 

\section{Two-stage vs.\ single-stage training comparison}
\label{sec:app-staging}

\begin{table}[b]
\caption{Two-stage vs.\ single-stage SFT across $42$ storefronts. Goal reached = successful task completion. Steps-limit = step budget exhausted. Other = malformed or unparseable agent output.}
\label{tab:staging-ablation}
 \vspace{5pt}
\centering
\begin{tabular}{lrrrr}
\toprule
& \multicolumn{2}{c}{\textbf{Two-Stage}} & \multicolumn{2}{c}{\textbf{Single-Stage}} \\
\cmidrule(lr){2-3} \cmidrule(lr){4-5}
\textbf{Outcome} & Count & \% & Count & \% \\
\midrule
Goal reached         & 42{,}083 & 83.5 & 41{,}720 & 82.8 \\
Steps-limit errors   &  6{,}625 & 13.1 &  6{,}333 & 12.6 \\
Time-limit errors    &     598 &  1.2 &       7 &  0.0 \\
Other errors         &  1{,}094 &  2.2 &  2{,}340 &  4.6 \\
\bottomrule
\end{tabular}
\end{table}

A central design choice in \textsc{SimPersona} is the two-stage SFT framework described in \Cref{sec:sft}. In Stage~1, we freeze the LLM backbone and train only the $256$ persona-token embeddings, and in Stage~2, we unfreeze the full model and jointly fine-tune all parameters. To test whether this decomposition is necessary, we compare against a single-stage baseline that trains all parameters jointly from initialization. Both variants use the same architecture, data, and optimization
settings; the only difference is whether persona grounding is decoupled from action learning.

We evaluate both models on the same simulation setup used in \Cref{sec:results-alignment}. Each simulation run has one of four terminal outcomes: \emph{goal reached}, \emph{steps-limit}, \emph{time-limit}, or \emph{other}. A steps-limit outcome means that the agent continued to produce valid actions but did not complete the assigned task within the $30$-step budget. A time-limit outcome means that the run exceeded the wall-clock timeout. The \emph{other} category captures malformed or non-executable outputs, including invalid JSON and structurally broken action strings. \Cref{tab:staging-ablation} reports aggregate outcome counts, while \Cref{tab:staging-stats} summarizes the distribution of per-shop error rates.
\begin{figure}[t]
  \centering
  \includegraphics[width=0.9\textwidth]{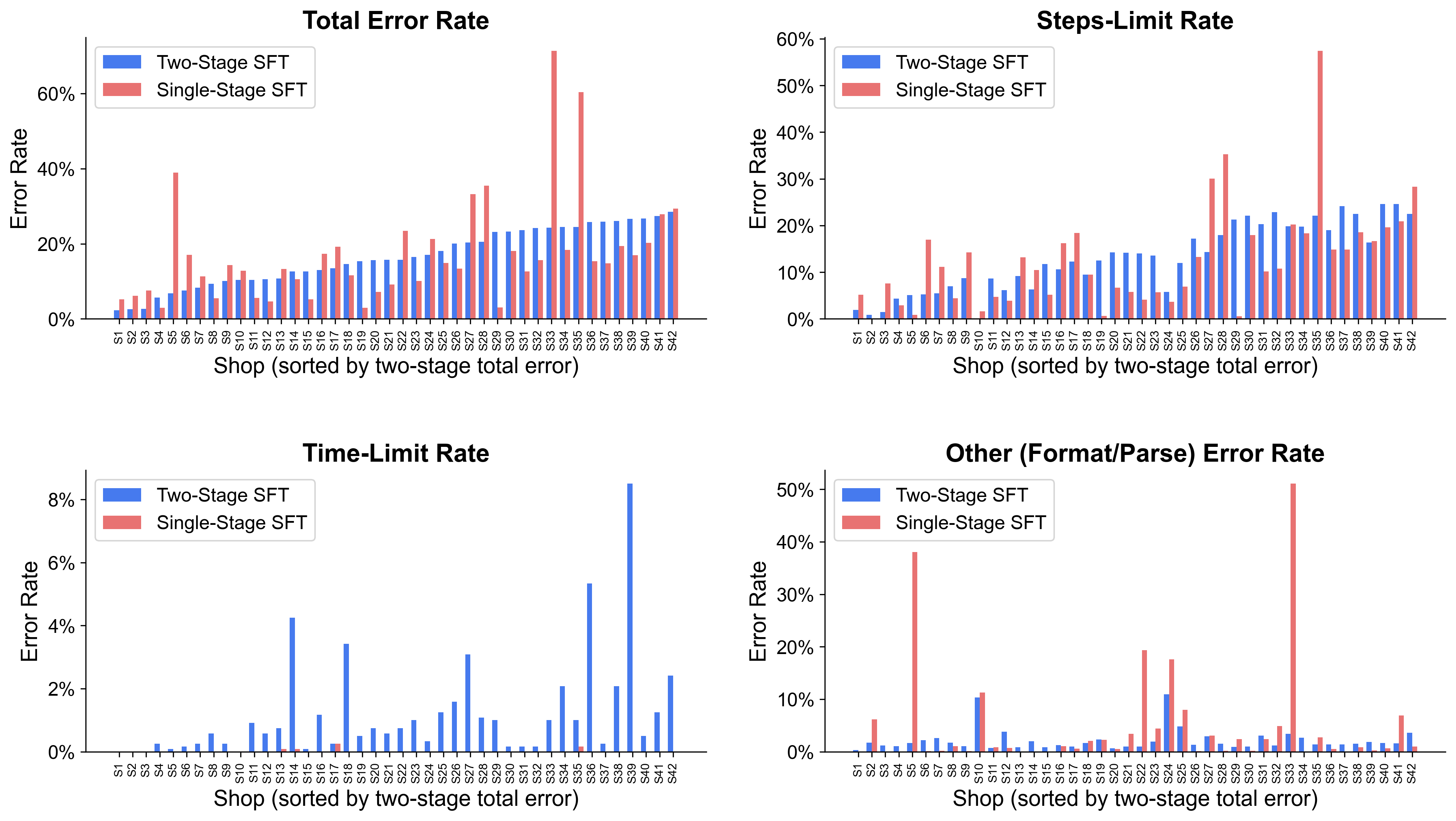}
  \caption{Per-shop error-rate comparison between two-stage and single-stage SFT (sorted by two-stage total error). The two-stage model keeps error bounded below $29\%$ on every shop, while single-stage training produces catastrophic outliers exceeding $70\%$. The non-zero time-limit rate under two-stage training reflects deeper storefront engagement rather than inefficiency.}
  \label{fig:staging-ablation}
\end{figure}

The aggregate goal-reached rates ($83.5\%$ vs.\ $82.8\%$) and overall error rates ($16.5\%$ vs.\ $17.2\%$) appear nearly identical, and a paired $t$-test on per-shop error rates confirms no significant difference in the mean
(p-value $= 0.72$).  However, focusing on the mean is misleading. The single-stage model's comparable average is an artifact of compensating extremes: it achieves very low error on some storefronts while collapsing catastrophically on others. The coefficient of variation, which measures how dispersed per-shop error rates are relative to their average, nearly doubles from $0.46$ (two-stage) to $0.81$ (single-stage). A coefficient of $0.81$ indicates that the typical deviation from the mean is almost as large as the mean itself, meaning that the aggregate error rate is a poor predictor of what any individual storefront will experience. The per-shop error range tells the same story: under two-stage training, the gap between the best and worst shop is $26.3$\,pp ($2.2\%$--$28.5\%$), whereas under single-stage training it balloons to $68.4$\,pp ($2.9\%$--$71.3\%$), a $2.6{\times}$ wider spread. In practical terms, this means a merchant deploying the single-stage model has no reliable guarantee of agent quality: while some storefronts enjoy error rates below $5\%$, others see more than half of all simulation episodes fail. As \Cref{tab:staging-stats} shows, our two-stage training framework does not primarily improve the mean, it eliminates tail risk, compress the error distribution so that no storefront exceeds $30\%$ error and every shop maintains at least $71.5\%$ goal completion.

\Cref{fig:staging-ablation} makes this distributional difference visually concrete. The two-stage error bars rise smoothly across shops and stay below $29\%$, while single-stage bars spike on several storefronts, revealing two distinct catastrophic failure modes. The first is \emph{structural output collapse}: on the worst shop, $51.1\%$ of episodes produce unparseable output, driving the total error to $71.3\%$; under two-stage training the same shop drops to $24.2\%$ error with only $3.4\%$ other errors; the entire $47$\,pp improvement comes from eliminating malformed outputs. The second mode is \emph{degenerate navigation}: another shop reaches $60.3\%$ error with $57.4\%$ of episodes exhausting the step budget in repetitive loops; two-stage training reduces this to $24.5\%$ by enabling more efficient action planning.

These worst cases reflect a systematic pattern. Total \emph{other} errors drop $53\%$ (from $2{,}340$ to $1{,}094$), and the per-shop other-error rate falls from $4.6\%$ mean (std $10.0$\,pp, max $51.1\%$) to $2.2\%$ mean (std $2.1$\,pp, max $10.9\%$). Across shops, the reduction in total error correlates strongly with the reduction in other errors (Pearson $r = 0.69$, $p < 10^{-6}$), confirming that structural output failures are the dominant source of variance eliminated by two-stage training. We attribute this to optimization interference: when $256$ randomly initialized persona-token embeddings are updated jointly with all backbone parameters, the large, initially noisy gradients can destabilize the structured-generation pathways responsible for producing valid JSON actions before the tokens acquire meaningful signal. The two-stage framework avoids this by freezing the backbone in Stage~1, letting the embeddings converge to coherent representations without corrupting action generation, so that Stage~2 fine-tuning starts from an aligned rather than conflicting initialization.

The remaining error-type panels reinforce this interpretation. Steps-limit rates are comparable in aggregate ($13.1\%$ vs.\ $12.6\%$) but far more variable under single-stage training (std $10.8$\,pp vs.\ $7.3$\,pp; one shop reaches $57.4\%$, while no two-stage shop exceeds $25\%$). The time-limit panel shows a counterintuitive asymmetry: the two-stage model incurs $1.2\%$ mean time-limit rate, while single-stage reports near-zero. The time-limit errors require the agent to actively render pages and interact  with DOM elements long enough to exhaust the wall-clock budget. The single-stage model on its hardest shops fails before reaching this regime, either through unparseable output or step-count exhaustion. The two-stage model engages deeply enough that it occasionally runs out of time while making genuine progress, making the presence of time-limit errors a proxy for \emph{behavioral depth}.
\section{Two stage training details} \label{sec:app-sft-examples}
\begin{figure}[h]
  \centering
  \begin{subfigure}{\textwidth}
    \centering
    \includegraphics[width=0.8\textwidth, trim=0 80 0 0, clip]{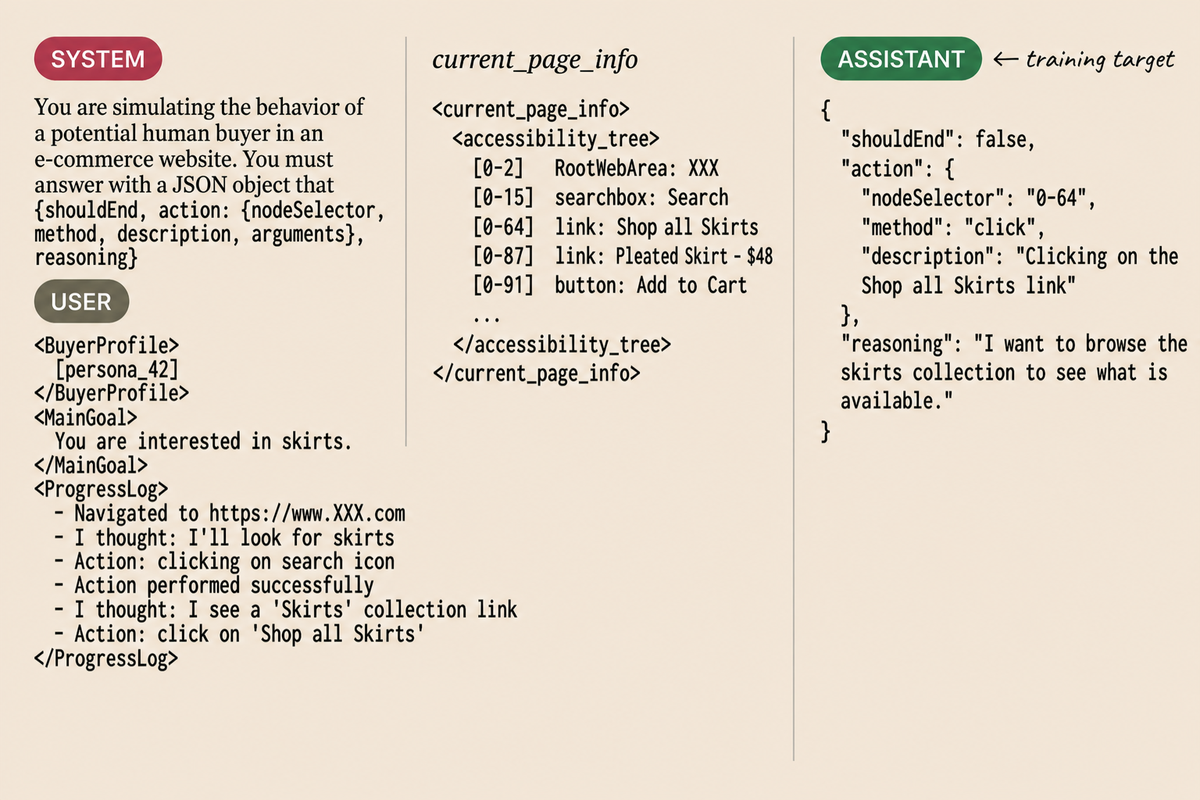}
    \caption{Stage~1: persona grounding (backbone frozen, only token embeddings updated).}
    \label{fig:sft-stage1}
  \end{subfigure}

  \vspace{10pt}

  \begin{subfigure}{\textwidth}
    \centering
    \includegraphics[width=0.8\textwidth]{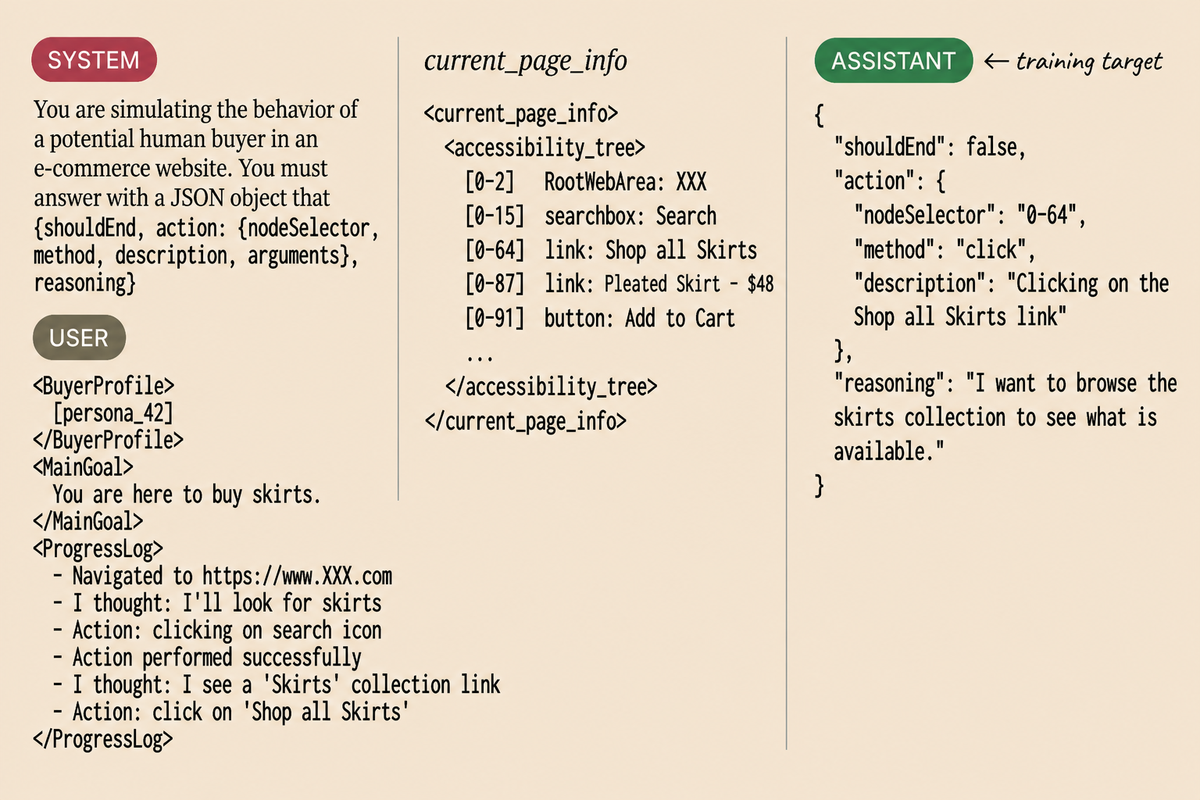}
    \caption{Stage~2: action-oriented fine-tuning (all parameters updated).}
    \label{fig:sft-stage2}
  \end{subfigure}
  \caption{Two-stage persona-grounding SFT examples. Each training example consists of a system prompt specifying the output schema, a user turn containing the persona token, session goal, progress log (memory), and current page DOM, and an assistant turn with a structured JSON action and reasoning trace. Stage~1 teaches the model \emph{what} each token means; Stage~2 teaches it \emph{how} to act on that knowledge.}
  \label{fig:sft-examples}
\end{figure}
Both stages use Qwen3-14B as the base architecture, trained on $2$ nodes $\times$ $8$ NVIDIA H200 GPUs ($16$ GPUs total) with Fully Sharded Data Parallel (FSDP).
In Stage~1, the backbone is fully frozen; only the $256$ persona-token embeddings are trainable.
We train for $1$ epoch on approximately $34{,}000$ multi-turn traces generated from $3{,}600$ sessions across $39$ shops (each session is on average $10$ steps), with an effective batch size of $64$ ($1$ per device $\times$ $16$ GPUs $\times$ $4$ gradient-accumulation steps), learning rate $1 \times 10^{-5}$ with cosine annealing and $50$ warmup steps, weight decay $0.01$, and maximum sequence length $16{,}384$ tokens. 
Loss is computed on assistant turns only.
Stage~2 follows the same configuration except that all model parameters are unfrozen and trained.
The $39$ training shops are fully disjoint from the $42$ evaluation storefronts used in \Cref{sec:results-alignment,sec:results-heads,sec:results-instruction}; no buyer appearing in the training set is included in any evaluation. Also, \Cref{fig:sft-examples} shows a schematic example of the data used for our two stage training described in \Cref{sec:sft}.

\section{Calinski--Harabasz index.} \label{sec:ch}
The Calinski--Harabasz (CH) index~\citet{calinski1974dendrite} is defined as
\begin{equation}
    \mathrm{CH}
    =
    \frac{\mathrm{SS}_{\mathrm{between}}\,/\,(K-1)}
         {\mathrm{SS}_{\mathrm{within}}\,/\,(N-K)},
\end{equation}
where $K$ is the number of clusters, $256$ in our case, $N$ is the total number of buyers, $\mathrm{SS}_{\mathrm{between}}$ is the sum of squared distances from each cluster centroid to the global centroid weighted by cluster size, and $\mathrm{SS}_{\mathrm{within}}$ is the sum of squared distances from each buyer to its assigned cluster centroid. Higher values indicate tighter, better-separated clusters.

\section{Persona token ablation under neutral intents}
\label{sec:app-token-ablation}

To further test the effect of persona tokens on agent behavior, we compare
simulations with persona-token conditioning against simulations without.
To ensure that the agent is not guided by intent-specific information, both
conditions use a neutral, generic intent (``you are interested in
product~X'').  The rest of the experimental setup is identical to
\Cref{sec:results-alignment}.

\begin{figure}[b]
  \centering
  \includegraphics[width=0.9\textwidth]{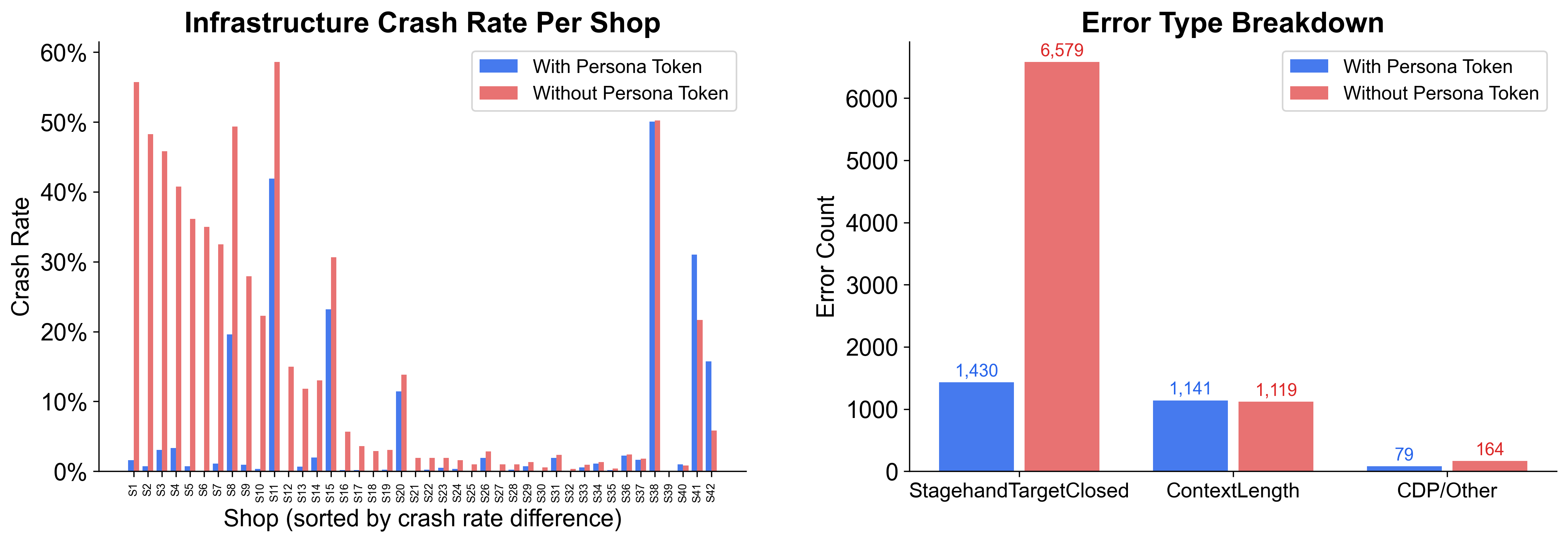}
  \caption{Persona token ablation under neutral intents. \textbf{Left}: per-shop infrastructure crash rate (sorted by difference); $38$ of $42$ shops crash more frequently without persona tokens. \textbf{Right}: error-type breakdown showing that browser crashes (\texttt{StagehandTargetClosed}) are $4.6{\times}$ more frequent without tokens, while context-length errors remain identical (${\approx}1{,}130$), isolating navigation behavior as the mechanism.}
  \label{fig:token-ablation}
\end{figure}

Our analysis reveals that removing persona tokens triples the rate of simulation crashes: $7{,}862$ ($15.6\%$) without tokens versus $2{,}650$ ($5.3\%$) with tokens (\Cref{tab:token-ablation}).  The gap is dominated by a single error category: \texttt{StagehandTargetClosedError}, a browser-level failure in which the page session terminates
mid-interaction.  This error occurs $4.6{\times}$ more frequently without persona tokens ($6{,}579$ vs.\ $1{,}430$).  To verify that this gap reflects a difference in navigation behavior rather than a difference in output formatting, we use context-length errors as a control: these errors are triggered when the model's prompt or response exceeds the token limit, a failure mode that depends entirely on output length and is independent of how the agent navigates the storefront.  Context-length errors are virtually identical across conditions ($1{,}119$ vs.\ $1{,}141$; ratio $0.98$), confirming that persona tokens do not change \emph{what} the agents generate but rather \emph{how} they interact with the page.  The crash reduction is not confined to a few outlier shops: $38$ of $42$ storefronts exhibit a higher crash rate without tokens, with per-shop differences exceeding $40$\,pp on the most affected shops (\Cref{fig:token-ablation}, left).

\begin{table}[b]
\caption{Distributional statistics of per-shop error rate across $42$ storefronts.}
\label{tab:staging-stats}
 \vspace{5pt}
\centering
\begin{tabular}{lcc}
\toprule
\textbf{Statistic} & \textbf{Two-Stage} & \textbf{Single-Stage} \\
\midrule
Std of error rate          & 7.6pp  & 13.9pp \\
Min error rate             & 2.2\%  & 2.9\%  \\
Max error rate             & 28.5\% & 71.3\% \\
\midrule
Shops $<$10\% error        & 8      & 13     \\
Shops $>$30\% error        & 0      & 5      \\
Shops $>$50\% error        & 0      & 2      \\
\midrule
Min goal-reached rate      & 71.5\% & 28.7\% \\
Shops $<$50\% goal reached & 0      & 2      \\
Shops $<$70\% goal reached & 0      & 5      \\
\bottomrule
\end{tabular}
\end{table}

\begin{table}[t]
\caption{Persona token ablation under neutral intents.  Persona tokens reduce browser crashes
  by $66\%$ and produce deeper storefront engagement across all behavioral
  metrics.}
\label{tab:token-ablation}
 \vspace{5pt}
\centering
\begin{tabular}{lccc}
\toprule
\textbf{Metric} & \textbf{With Token} & \textbf{No Token} & \textbf{Ratio} \\
\midrule
\multicolumn{4}{l}{\textit{Episode outcomes (all simulations)}} \\
\quad Simulation crashes         & $2{,}650$ ($5.3\%$)   & $7{,}862$ ($15.6\%$)  & $0.34{\times}$ \\
\quad\quad StagehandTargetClosed & $1{,}430$             & $6{,}579$             & $0.22{\times}$ \\
\quad\quad Context length        & $1{,}141$             & $1{,}119$             & $1.02{\times}$ \\
\quad Goal reached               & $39{,}664$ ($78.7\%$) & $37{,}942$ ($75.3\%$) & $1.05{\times}$ \\
\midrule
\multicolumn{4}{l}{\textit{Behavioral engagement (all simulations)}} \\
\quad Add-to-cart events         & $16{,}132$            & $13{,}698$            & $1.18{\times}$ \\
\quad Checkout reached           & $7{,}823$             & $6{,}664$             & $1.17{\times}$ \\
\midrule
\multicolumn{4}{l}{\textit{Navigation depth (succeeded simulations only)}} \\
\quad Avg actions / session      & $11.1$                & $9.1$                 & $1.23{\times}$ \\
\quad Avg pages visited          & $3.9$                 & $3.6$                 & $1.10{\times}$ \\
\quad Avg products viewed        & $1.39$                & $1.29$                & $1.08{\times}$ \\
\midrule
\multicolumn{4}{l}{\textit{Crash rate consistency}} \\
\quad Shops w/ higher crash rate & $3/42$                & $38/42$               & --- \\
\bottomrule
\end{tabular}
\end{table}

\texttt{StagehandTargetClosedError} occurs when the agent's actions destabilize the browser session, for example, by clicking elements that no longer exist after a page transition, issuing navigation commands faster than the renderer can process, or interacting with the DOM during partial page loads.  These failures are deterministic
consequences of the agent's action sequence, not stochastic infrastructure events.  In our simulations, both conditions receive only a neutral intent (``you are interested in product~X'') that specifies a product but provides no behavioral signal. Without a persona token, the agent lacks any prior over trajectory structure and it must select actions conditioned solely on page observations and a generic goal. This yields a high-entropy action distribution that produces erratic navigation like frequent page switches, interactions with transient UI elements, and action--observation desynchronization during asynchronous page loads.  These are precisely the trajectory patterns that maximize the probability of acting on stale page state. A persona token provides the missing behavioral signal.  Trained on real buyer navigation traces, each token encodes a coherent browsing style that constrains the agent's action distribution toward structured, sequential trajectories.  Because storefronts are designed for human navigation, these human-like trajectories are inherently more compatible with the browser environment, explaining why the token reduces \texttt{StagehandTargetClosed} crashes by $78\%$ without affecting context-length errors at all.

Behavioral engagement metrics provide further evidence that persona tokens produce deeper, more purposeful navigation rather than merely preventing crashes (\Cref{tab:token-ablation}).  Across all simulations, persona-conditioned agents add items to cart $18\%$ more often ($16{,}132$
vs.\ $13{,}698$) and reach checkout $17\%$ more frequently ($7{,}823$ vs.\ $6{,}664$).  On clean (non-crash) simulations where both conditions complete the session without infrastructure failure, persona-conditioned agents
still perform $23\%$ more actions per session ($11.1$ vs.\ $9.1$), visit $10\%$ more pages ($3.9$ vs.\ $3.6$), and view $8\%$ more products ($1.39$ vs.\ $1.29$).  The persona token therefore provides a genuine behavioral prior that shapes navigation even in the absence of persona-specific intent: the agent browses more deeply, interacts with more product pages, and progresses further through the purchase funnel, producing sessions that more closely resemble real buyer behavior.
\section{Instruction-following task diversity}
\label{sec:app-tasks}
To evaluate instruction-following quality (\Cref{sec:results-instruction}), we construct a deterministic benchmark of navigation tasks with varying complexity. Each task specifies a concrete sequence of actions the agent must execute on a live storefront e.g., searching for specific products, viewing product pages, adding items to cart, and proceeding through checkout. \Cref{tab:task-diversity} shows representative examples ordered by complexity: the simplest tasks require 3 actions (search, view, cart), while the most complex involve 27 actions spanning multiple search queries, product comparisons, cart modifications, and a completed purchase. 

\begin{table}[h]
\caption{Representative tasks from the deterministic benchmark, ordered by complexity.  Bins denote (Engagement\,/\,Exploration\,/\,Purchase); \emph{Nav.}\ counts distinct navigation actions in the instruction.}
\label{tab:task-diversity}
 \vspace{5pt}
\centering
\begin{tabular}{ccl}
\toprule
\textbf{Bins (E/X/P)} & \textbf{Nav.} & \textbf{Task description} \\
\midrule
0\,/\,2\,/\,1 & 3  & Search for one product, view it, add to cart, end session \\
0\,/\,1\,/\,1 & 5  & Search for one product, view it, add to cart, abandon checkout \\
1\,/\,2\,/\,0 & 7  & Search two products, view each, add one to cart, abandon checkout \\
2\,/\,2\,/\,1 & 10 & Search 3 products, view each, add one, revisit pages, abandon \\
2\,/\,2\,/\,2 & 16 & View 4 products, add one to cart, view 2 more, add another, checkout \\
2\,/\,2\,/\,2 & 27 & Search and view 6 products, add 3 to cart, revisit 2, purchase \\
\bottomrule
\end{tabular}
\end{table}


\end{document}